\documentclass[sigconf]{acmart}

\usepackage{amsmath}
\usepackage{xurl}
\usepackage{multirow}
\usepackage{mathtools}
\usepackage{booktabs}
\usepackage{subfig}
\usepackage{graphicx}
\usepackage[utf8]{inputenc}
\usepackage[para,online,flushleft]{threeparttable}
\usepackage{sidecap}
\usepackage{tablefootnote}
\usepackage{tikz}
\usepackage{soul}
\usepackage{url}

\newcommand*\circled[1]{\tikz[baseline=(char.base)]{
            \node[shape=circle,fill,inner sep=1pt] (char) {\footnotesize \textcolor{white}{#1}};}}

\newcommand{\gnn}{\textsc{Tango}}

\begin{document}

\date{}

\title{
{\gnn}: rethinking quantization for\\
graph neural network training on GPUs
}
\author{\fontsize{10}{12}\selectfont Shiyang Chen}\affiliation{\fontsize{8}{10}\selectfont \institution{Rutgers, The State University of New Jersey}
\country{}}

\author{\fontsize{10}{12}\selectfont Da Zheng}
\authornote{The work is not related to the author's position at Amazon.}
\affiliation{\fontsize{8}{10}\selectfont
	\institution{Amazon}
	\country{}
}

\author{\fontsize{10}{12}\selectfont  Caiwen Ding}
\affiliation{\fontsize{8}{10}\selectfont
	\institution{ University of Connecticut}
	\country{}}

\author{\fontsize{10}{12}\selectfont Chengying Huan}
\affiliation{\fontsize{8}{10}\selectfont
	\institution{Institution of Software, Chinese Academy of Sciences}
	\country{}
}

\author{\fontsize{10}{12}\selectfont Yuede Ji}
\affiliation{\fontsize{8}{10}\selectfont
	\institution{University of North Texas}
	\country{}}
	
\author{\fontsize{10}{12}\selectfont Hang Liu}
\affiliation{\fontsize{8}{10}\selectfont
	\institution{Rutgers, The State University of New Jersey}
	\country{}}




\copyrightyear{2023}
\acmYear{2023}
\setcopyright{acmlicensed}\acmConference[SC '23]{The International Conference for High Performance Computing, Networking, Storage and Analysis}{November 12--17, 2023}{Denver, CO, USA}
\acmBooktitle{The International Conference for High Performance Computing, Networking, Storage and Analysis (SC '23), November 12--17, 2023, Denver, CO, USA}
\acmPrice{15.00}
\acmDOI{10.1145/3581784.3607037}
\acmISBN{979-8-4007-0109-2/23/11}

\thispagestyle{empty}

%
\begin{abstract}
Graph learning is becoming increasingly popular due to its superior performance in tackling many grand challenges.
While quantization is widely used to accelerate Graph Neural Network (GNN) computation, quantized training faces remarkable roadblocks. 
Current quantized GNN training systems often experience longer training time than their full-precision counterparts for two reasons: (i) addressing the quantization accuracy challenge leads to excessive overhead, and (ii) the optimization potential exposed by quantization is not adequately leveraged.
This paper introduces {\gnn} which re-\ul{t}hinks qu\ul{an}tization \textit{challenges} and \textit{opportunities} for \ul{g}raph neural netw\ul{o}rk training on GPUs with three contributions: 
Firstly, we introduce efficient rules to maintain accuracy during quantized GNN training. Secondly, we design and implement quantization-aware primitives and inter-primitive optimizations to speed up GNN training. Finally, we integrate {\gnn} with the popular Deep Graph Library (DGL) system and demonstrate its superior performance over the state-of-the-art approaches on various GNN models and datasets.

\end{abstract}
\maketitle
%

\begin{figure*}[h!t]
\centering
\subfloat[GAT forward propagation.]{
\hspace{-.05in}
\includegraphics[width=.48\linewidth]{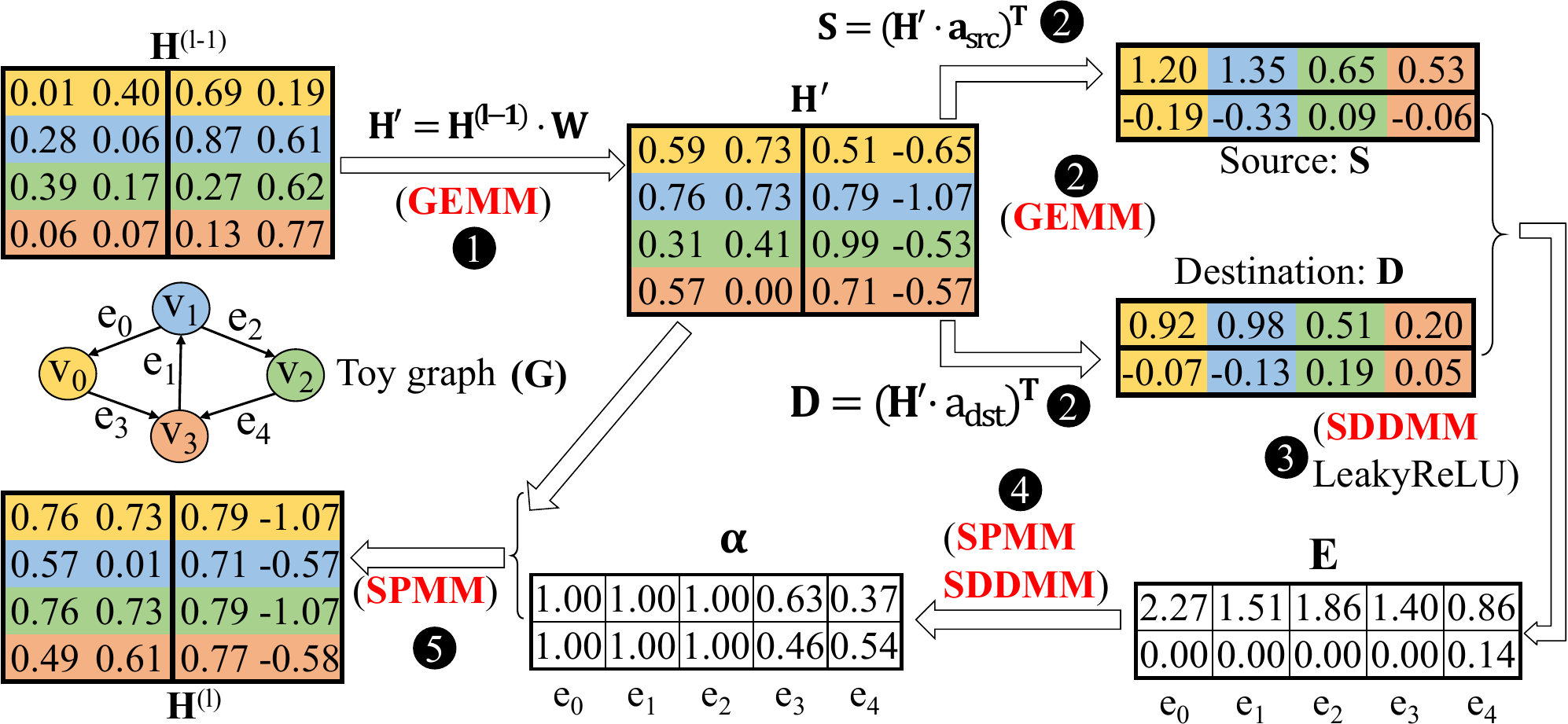}
\label{fig:GAT_forward}
}
\quad
\subfloat[GAT backward propagation.]{
\hspace{-.05in}
\includegraphics[width=.48\linewidth]{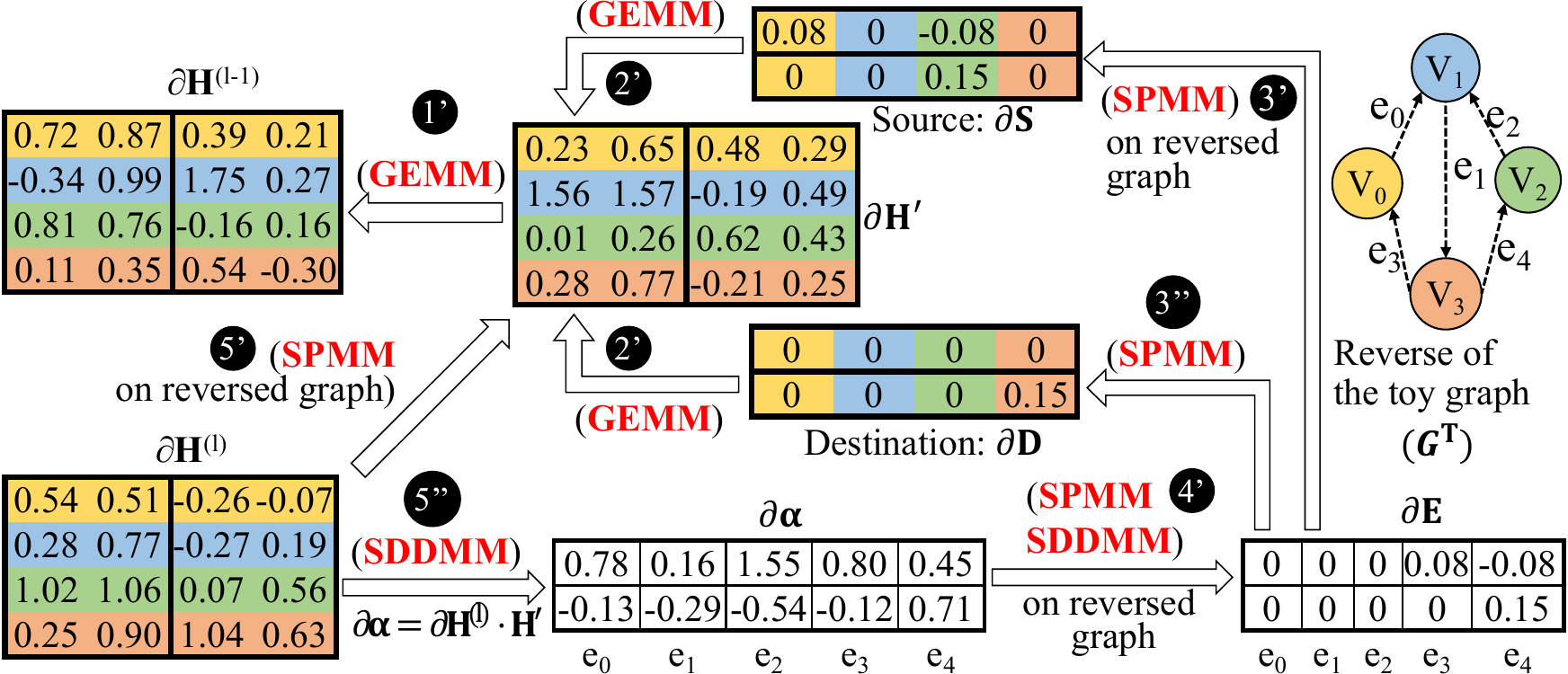}
\label{fig:GAT_back}
}
\caption{GAT training on a toy graph, i.e., middle left of (a), with two heads. This example is used throughout this paper.}
\label{fig:GAT}
\end{figure*}

\section{Introduction}\label{sec:intr}

Graph analytics can claim a large share of the credit for tackling many grand challenges of our time -- such as understanding the spread of pandemics~\cite{marathe2013computational}, designing extremely large-scale integrated circuits~\cite{zhang2019circuit}, and uncovering software vulnerabilities~\cite{xu2017neural}, among many others~\cite{gaihre2021dr,banerjee20013,FINGERS,chen2021re,gaihre2019xbfs, liu2019simd,GraphP,lam2022graphcast}.
In particular, since the introduction of the Graph Convolution Network (GCN) by Kipf and Welling in 2016~\cite{kipf2016semi}, GNNs have gained widespread popularity as a vibrant field in graph analytics.
{This is primarily because various GNN models have shown promising results in addressing these challenges through node and edge embedding-based designs.}~\cite{hamilton2017inductive,NIPS2013_1cecc7a7,zheng2020dgl,InformationNidhi2021,xu2022payment,lim2021visual,informatics10010008,won2022ulppack,GraSP}.

\vspace{.05in}
A typical GNN model often performs \textit{linear transformation}, and \textit{graph structure related operations} on node feature (\textbf{H}), edge feature (\textbf{E}), and graph structure (\textbf{G}). Using Graph Attention Network (GAT)~\cite{GAT2018graph} (refer to Figure~\ref{fig:GAT}, next page) as an example, this model (i) applies a multilayer perception on node feature matrix, (ii) uses graph topology to derive the edge features, and (iii) calculates the destination feature by considering both the source node and edge features. One could further extend the aforementioned steps (i) - (iii) to multi-hop neighbors to derive multi-layer GATs. 
Of note, step (i) is a GEneral Matrix Multiply (GEMM) primitive, while steps (ii) and (iii) are sparse primitives, i.e., SParse-dense Matrix Multiplication (SPMM) and Sampled Dense-Dense Matrix product (SDDMM), whose sparse nature is usually defined by the graph. 



%
\vspace{.05in}
Quantization is a primary approach to accelerating Deep Neural Network (DNN) and GNN models for two major optimizations \textit{opportunities}, that is, quantization could lead to both computation and data access reductions. 
{On the one hand, for computation-intensive primitives, e.g., GEMM, computing on quantized data is faster than on a floating-point counterpart. For instance, computing with 8-bit integers on tensor core offers 2$\times$ the throughput of 16-bit floating-point and 32$\times$ that of 32-bit floating-point, respectively~\cite{tensorcore}.
}
On the other hand, for sparse primitives, i.e., SPMM and SDDMM, which are data-intensive, quantization {reduces the size of tensors}, thus reducing memory traffic and time consumption.

\vspace{.05in}
Whereas the \textit{challenge} is that quantization errors (i.e., caused by fewer bits) could prevent the model from achieving the desired accuracy.
Correspondingly, there mainly exist three tracks of research efforts: (i)
For multiply-and-accumulate operations in matrix multiplication, limited precision can cause values of different magnitudes to accumulate inaccurately. The proposed solutions are chunk-based accumulation~\cite{wang2018training} and dynamically adjustable data formats~\cite{koster2017flexpoint,das2018mixed,sakr2018per}.
(ii) For weight updates in backpropagation, quantization errors could negate the gradient update to the weights. 
SWALP~\cite{yang2019swalp} proposes accumulating and updating the weight after multiple training epochs.  
(iii) 
To mitigate divergence from quantization errors, Zhu et al.~\cite{zhu2020towards} propose heuristics for gradient clipping and learning rate adjustments.

\vspace{.05in}
Contemporary quantized GNN training systems often experience longer training time than their full-precision counterparts for two reasons: 
\textit{(i) addressing the accuracy challenge results in significant overhead.}
The computation of the proposed novel data formats is inefficient because commodity GPUs support neither fixed-point data format nor floating-point format with dynamically adjusted exponent and mantissa. 
The model trained using SWALP or clipped gradients needs more epochs to converge because the weights are updated less frequently. 
\textit{(ii) The optimization potential offered by quantization is not well-utilized. }
ActNN~\cite{chen2021actnn}, TinyKG~\cite{chen2022tinykg}, and EXACT~\cite{liu2022exact} quantize the tensors to save memory and dequantize them back to full-precision for computation, increasing the overall training time~\cite{liu2022gact,evans2021ac}. For example, TinyKG with 8-bit quantization is 54.1\% slower than using FP32. Of note, Degree-Quant~\cite{tailor2021degreequant} performs Quantization-Aware Training (QAT)~\cite{zafrir2019q8bert,bhandare2019efficient,jin2020adabits,dong2019hawq,esser2015backpropagation,courbariaux2015binaryconnect,jacob2018quantization} that uses full-precision to ``simulate quantization'' in training to reduce the error for quantized inference, which, again, experiences longer training time.

\vspace{.05in}
This paper introduces {\gnn},
the first GPU-based quantized GNN training system that both \textit{maintains the model accuracy} and \textit{reduces turnaround time} when compared to the full precision counterpart. 
Particularly, {\gnn} encompasses three contributions: 

\begin{itemize}
    \item We introduce several lightweight rules to maintain accuracy for quantized GNN training. The rules include GPU-accelerated stochastic rounding, derivation of proper quantization bit count, novel quantization-aware GEMM, SPMM, and SDDMM, and full precision weight update and softmax. 
\vspace{.05in}

    \item We design and implement a quantization-aware system to reduce the GNN training time on GPUs. Our techniques include GEMM with on-the-fly quantization, incidence-matrix-based adaptive SPMM,  SDDMM with on-the-fly dequantization, and inter-primitive optimizations.
\vspace{.05in}

    \item For ease of use, we integrate {\gnn} with DGL, which uses PyTorch as the backend. Therefore, all existing DGL-based models can enjoy the performance benefits from {\gnn} without any changes. We demonstrate that {\gnn} constantly outperforms state of the art for all evaluated GNN models while maintaining the training accuracy. 
\end{itemize}

The remainder of this paper is organized as follows: Section~\ref{sec:back} presents the background. Section~\ref{sec:framework} discusses the design and techniques in {\gnn}. Specifically, Section~\ref{sec:overview} describes the challenges and opportunities of {\gnn}, Section~\ref{sec:acc} illustrates the lightweight rules for maintaining training accuracy during quantization, and Section~\ref{subsec:fast} presents the systematic effort on quantization accelerated training.  
We evaluate {\gnn} in Section~\ref{sec:eval}, describe related work in Section~\ref{sec:related}, and conclude in Section~\ref{sec:conclude}.

\section{Background}\label{sec:back}

\subsection{A running example for GAT training}\label{subsec:primitive}

This section uses a running example to illustrate how to express the forward and backward computations of GAT with three key primitives, i.e., GEMM, SPMM, and SDDMM.  

\vspace{.05in}\noindent
\textbf{Primitives for forward computation.}
Figure~\ref{fig:GAT_forward} presents the forward workflow of GAT on a toy graph, i.e., {node projection} (\circled{1} - \circled{2}), {attention computation} (\circled{3} - \circled{4}), and {message aggregation} (\circled{5}). 

In step \circled{1}, 
GAT resorts to \textit{GEMM primitive} to perform a linear transformation for node features, i.e., \textbf{H}$'$ = \textbf{H}$^{(l-1)}\cdot \mathbf{W}$. In \textbf{H}$^{(l-1)}$, where each row of \textbf{H}$^{(l-1)}$, in the same color, represents the features of one node. Each node feature contains two heads in \textbf{H}$^{(l-1)}$. Using the first row of \textbf{H}$^{(l-1)}$ as an example, 
[0.01, 0.40] is the first head, and the remaining two values belong to the second head. \textbf{W} is the learnable weight matrix for linear transformation.

In step \circled{2}, GAT consolidates each head of the feature vector into one scalar by GEMM,  
i.e., $\textbf{S}=(\textbf{H}'\cdot\textbf{a}_{src})^T$ and $\textbf{D}=(\textbf{H}'\cdot\textbf{a}_{dst})^T$. 
Using node $\mathbf{v}_0$ of \circled{2} as an example, 
for the source feature, [0.59, 0.73]$\times$[0.91, 0.90]$^T$ = 1.20, and [0.51, -0.65]$\times$[0.42, 0.62] = -0.19. Similarly, we can derive the entire \textbf{S} and \textbf{D}.


In step \circled{3}, the source {(\textbf{S})} and destination {(\textbf{D})} node feature matrices are combined by an \textit{SDDMM primitive} to arrive at the edge feature \textbf{E}. 
Formally, the SDDMM is defined as $\mathbf{E} = \mathbf{G} \odot(\mathbf{S} \oplus \mathbf{D}^T)$, where 
every row of \textbf{S} computes against every row of \textbf{D} with the customized operation $\oplus$, and $\odot$ is a Hadamard product operator. The resultant matrix \textbf{E} is masked out by the sparse adjacency matrix \textbf{G}
of the graph so $\textbf{E}[i][j] = 0$ when there is no edge between $v_i$ and $v_j$.
In Figure~\ref{fig:GAT_forward}, $\oplus$ denotes addition. Using edge $e_3$ as an example, since it connects source $v_0$ and destination $v_3$, we arrive at [1.20, -0.19] + [0.20, 0.05] = [1.40, -0.14] for $e_3$. Then an element-wise LeakyReLU is applied to the edge features. 
Particularly, each non-negative entry in \textbf{E} is unchanged while negative ones become close to $0$, which {we} use $0$ to represent. Hence [1.40, -0.14] becomes [1.40, 0.00] in Figure~\ref{fig:GAT_forward}.

In step \circled{4}, edges of the same destination come together to compute the head-wise attention scores through softmax operation. 
Using node $v_3$ as an example, it has $e_3$ and $e_4$ as the incoming edges. 
Therefore, the attention scores of $e_3$ and $e_4$ are computed as 0.63 = $\frac{e^{1.40}}{e^{1.40} + e^{0.86}}$, 
and 0.46 = $\frac{e^{0}}{e^{0} + e^{0.14}}$ for $e_3$, and 0.37 = $\frac{e^{0.86}}{e^{1.41} + e^{0.86}}$ and 0.54 = $\frac{e^{0.14}}{e^{0} + e^{0.14}}$ for $e_4$.
Putting this example into a general formula, we use SPMM and SDDMM operations together to compute the denominator as follows:
First, we use \textit{SPMM} to aggregate the in-edges for every node such as $\mathbf{M'} = (\mathbf{G}\odot exp(\mathbf{E})) \cdot \mathbf{1}$. Of note, $\mathbf{1}$ is an all `1' dense matrix. Second, since the first step computed the denominator for each destination vertex, we use \textit{SDDMM} to assign this denominator back to each incoming edge via $\mathbf{E'} = \mathbf{G} \odot(\mathbf{1} \cdot \mathbf{M'}^T)$. Subsequently, $\boldsymbol{\alpha} = \frac{exp(\mathbf{E})}{\mathbf{E'}}$.




In step \circled{5}, GAT performs an \textit{SPMM} to derive the new node embedding via 
\textbf{H}$^{(l)}$ = $(\mathbf{G}\odot \boldsymbol{\alpha}) \cdot \textbf{H}'$.
Intuitively, this step derives the new embedding by computing the weighted sum of all the incoming neighbors to a destination vertex. 
Using $v_3$ as an example, its incoming neighbors are \{v$_0$, v$_2$\}. We arrive at $\mathbf{H}^{(l)}[v_3]=\boldsymbol\alpha[e_3] \cdot \mathbf{H'}[v_0] + \boldsymbol\alpha[e_4] \cdot \mathbf{H'}[v_2]$, resulting in [0.49, 0.61, 0.77, -0.58].

\vspace{.05in}\noindent
\textbf{Primitives for backward computation.}
Figure~\ref{fig:GAT_back} is the backward pass of Figure~\ref{fig:GAT_forward}.
Steps \circled{5'} (\textit{SPMM}) and \circled{5''} (\textit{SDDMM}) of Figure~\ref{fig:GAT_back} are the corresponding backward operations for step \circled{5} in Figure~\ref{fig:GAT_forward}.
In the forward SPMM operation (\circled{5}), 
\textbf{H}$^{(l)}$ = 
$(\mathbf{G}\odot \boldsymbol{\alpha}) \cdot \textbf{H}'$,
we hence disperse the gradients of $\mathbf{H}^{(l)}$ to both $\mathbf{H}'$ and $\boldsymbol\alpha$.
First, we arrive at 
$\partial \mathbf{H}' = (\mathbf{G}^T\odot \boldsymbol{\alpha})\cdot\partial \mathbf{H}^{(l)}$
, which is an \textit{SPMM} (\circled{5'}) on the reversed graph since the updated node feature \textbf{H}$^{(l)}$ is aggregated from the source node feature.
Using $\partial\textbf{H}[v_1]$ as an example, it receives gradients from both $e_0$ and $e_2$. Therefore, we arrive at $\partial\mathbf{H}^{(l-1)}[v_1]=\boldsymbol\alpha[e_0] \cdot \partial\mathbf{H}^{(l)}[v_0] + \boldsymbol\alpha[e_2] \cdot \partial\mathbf{H}^{(l)}[v_2]$. That is, [1.56,1.57,-0.19,0.49]= 1.0$\times$[0.54, 0.51] + 1.0$\times$[1.02,1.06] $\|_{concat}$ 1.0$\times$[-0.26, -0.07] + 1.0$\times$[0.07,0.56]. 
Second, we have 
 $\partial\boldsymbol\alpha = \mathbf{G} \odot(\partial \mathbf{H}^{(l)} \cdot \textbf{H}'^T)$, 
which is an SDDMM operator on the original graph, where it performs row-wise dot-product (\circled{5''}). Using $\partial\boldsymbol\alpha[e_0]$ as an example, it connects nodes $v_1$ and $v_0$, we arrive at $\partial\boldsymbol\alpha[e_0]=\partial \mathbf{H}^{(l)}[v_0]\cdot \mathbf{H}'[v_1]$. In the example, we get 
[0.78,-0.13] =[0.54,0.51] $\times$ [0.76,0.73] $^T\|_{concat}$[-0.26,-0.07]$\times$[0.79,-1.07]$^T$.

Step \circled{4'} computes the gradient of edge features using attention scores. 
Using $e_3$ as an example, we compute $\partial \mathbf{E}[e_3]=\boldsymbol{ \alpha}[e_3](\partial\boldsymbol{ \alpha}[e_3]-(\partial\boldsymbol{\alpha}[e_3]\boldsymbol{ \alpha}[e_3]+\partial\boldsymbol{\alpha}[e_4]\boldsymbol{\alpha}[e_4]))$ based on the derivative of softmax operation.
We first use \textit{SPMM} to aggregate the incoming edge features for every node $\mathbf{P} = (\mathbf{G}\odot \partial\boldsymbol{\alpha} \odot \boldsymbol{\alpha}) \cdot \mathbf{1}$. For example, $\partial\boldsymbol{\alpha}[e_3]\boldsymbol{\alpha}[e_3]+\partial\boldsymbol{\alpha}[e_4]\boldsymbol{\alpha}[e_4]$ is the aggregation on $v_3$ as 0.80$\times$0.63 + 0.45$\times$0.37 = 0.67 for the first head.
Then we compute the final gradient for every edge with \textit{SDDMM}, 
$\partial\mathbf{E} = \boldsymbol{\alpha} 
\odot ( \partial\boldsymbol{\alpha} -(\mathbf{G}^T \odot(\mathbf{P} \cdot \mathbf{1}^T)))$.
That is, every node feature \textbf{P} is assigned to their out-edges in the reversed graph, and then computed with $\partial\boldsymbol{\alpha}$ and $\boldsymbol{\alpha}$. The gradient of the first head of $e_3$ is 0.63 $\times$ (0.80 - 0.67) = 0.08.





In step \circled{3'} and \circled{3''}, the gradient of edge attention score is used to compute the source feature and destination feature with two \textit{SPMM} operations, $\partial\textbf{S}=(\mathbf{G}^T \odot \partial\textbf{E}) \cdot \mathbf{1}$ and $\partial\textbf{D}=(\mathbf{G} \odot \partial\textbf{E}) \cdot \mathbf{1}$, where nodes aggregate their out-edge and in-edge attention scores, respectively. Still use $v_3$ as an example, its gradient $\partial\textbf{S}$[$v_3$] = $\partial\textbf{E}$[$e_1$] = [0, 0] and $\partial\textbf{D}$[$v_3$] = $\partial\textbf{E}$[$e_3$] + $\partial\textbf{E}$[$e_4$] = [0, 0.15]. The gradients from multiple out-edges are accumulated.


Note steps \circled{2'} and \circled{1'}, which do not depend on the graph structure, will follow the traditional DNN back propagation method for gradient computation. We skip the details.

\vspace{-.1in}\subsection{GNN models}
\label{subsec:back:gnn}
 

There exist a variety of GNN models. 
Graph Convolutional Network (GCN)~\cite{kipf2016semi} 
derives a graph convolutional operator through spectral graph theory. It can be expressed by GEMM and SPMM operations. 
%
Later, GraphSAGE~\cite{hamilton2017inductive} uses sampling to encode the graph topology for inductive learning. 
The model is applicable for unseen nodes because it learns the feature from sampled sub-graph.
\textit{GraphSAGE can be implemented with GEMM and SPMM. 
}
GAT~\cite{GAT2018graph} further introduces graph attention mechanisms that can attend to various neighbors with weights. 
This model contains GEMM, SPMM, and SDDMM primitives. 
%
Later, Relational GCN (RGCN) extends GCN via assigning different parameters to edges with different types~\cite{RGCN,7536145}. 
\textit{Here, RGCN consists of GEMM and SPMM primitives.} 
HGT proposes a transformer-based GNN model for heterogeneous graphs~\cite{hu2020heterogeneous}. It contains different parameters for distinct edge and node types. 
\textit{This model includes GEMM, SDDMM, and SPMM primitives.} 


We study two GNN models, i.e., GCN and GAT, for two reasons: (i) These two models are the most popular and cover all the required primitives for most GNN models. (ii) These two models contain relatively large and complete training and testing datasets.

\vspace{-.1in}
\subsection{Quantization}
\label{subsec:back:quant}

For a collection of values $\textbf{X}=\{\textbf{X}_i\ |\ \textbf{X}_i\in[\textbf{X}_{min}, \textbf{X}_{max}]\}$ which are represented in full precision, quantization uses fewer number of bits (i.e., \textit{B}) to represent each $\textbf{X}_i$. Quantization scatters \textbf{X} into $2^B-1$ buckets. Subsequently, all the \textbf{X}$_i$'s in the same bucket are represented as the same value, i.e., the bucket value.

For \textit{uniform} quantization, we assign each bucket the same value range, that is, $s=\frac{\alpha-\beta}{2^B-1}$, where $[\alpha,\beta]$ is the clipping range of \textbf{X}. There are also \textit{nonuniform quantization} whose quantized values are not necessarily uniformly spaced. If one wants to include the entire value range of \textbf{X}, one needs $\alpha=\textbf{X}_{min}$, and $\beta=\textbf{X}_{max}$. Formally, 
\begin{equation}
    \textbf{X}_{i, Quant} = round(\frac{\textbf{X}_i}{s}) - Z,
    \label{eq:back:quant}
\end{equation}
where $Z=\frac{\alpha+\beta}{2}$ is the zero point after quantization. One can recover the original value \textbf{X}$_i$ by dequantizing $\textbf{X}_{i,Quant}$:
\begin{equation}
    {\textbf{X}_i} \approx s\cdot(\textbf{X}_{i, Quant} + Z).
    \label{eq:back:dequant}
\end{equation}

Quantization further includes the following three configurations: (i) \textit{Asymmetric} vs. \textit{symmetric} quantization. Particularly, for $s$, one can let $-\alpha$=$\beta$=max($|\textbf{X}_{max}|$, $|\textbf{X}_{min}|$). While asymmetric quantization will likely enjoy a more precise clipping range when compared to symmetric quantization, the latter design, however, simplifies the quantization function in Equation~\ref{eq:back:quant} as $Z=\frac{\alpha+\beta}{2}=0$. (ii) \textit{Quantization granularity} concerns about the size of \textbf{X}. Using a matrix as an example, we can extract the same $s$ for the entire matrix or one $s$ per row/column of a matrix. The latter has a finer granularity than the former.
(iii) \textit{Static vs dynamic} quantization determines whether we change $s$ for the same tensor \textbf{X} from iteration to iteration. The dynamic version does so, while the static one does not. 
In {\gnn}, we adopt \textit{symmetric}, \textit{tensor-level granularity}, \textit{dynamic} quantization to maintain training accuracy and enhance training speed.

\section{{\gnn}: an accuracy and speed co-designed quantization system}\label{sec:framework}
\subsection{{\gnn} overview} \label{sec:overview}

Our key observation is that quantization presents both \textit{challenges} and \textit{opportunities} for GNN training. {\gnn} aims to tackle the challenges efficiently while extracting quantization benefits as follows: 



\vspace{.07in}

\textbf{Challenge: Maintaining training accuracy} poses three issues: (i) quantization could introduce additional computation tasks in addition to the steps in Figure~\ref{fig:GAT}. We need to reduce the overhead brought by those additional computations. (ii) For various operations in GNN training (in Figure~\ref{fig:GAT}), we need to decide what operations should be quantized and how we should quantize the tensors in each operator to meet the training accuracy requirements. In addition, (iii) those rules should expose optimization opportunities for {\gnn} to accelerate the most time-consuming operations in GNN training with quantization. 


In this paper, (i) we introduce GPU-friendly stochastic rounding and a lightweight operation to determine the required \# of quantization bits, reducing the cost of meeting accuracy requirements. (ii) We determine that weight update and softmax operations should be performed in full precision, while GEMM, SPMM, and SDDMM can be performed in our novel quantization-aware manner. This minimizes the impact on training accuracy while providing critical optimization opportunities for reducing turnaround time. Notably, (iii) GEMM, SPMM, and SDDMM are the most time-consuming phases in GNN, and our quantization-aware design offers optimization opportunities (see below) to reduce computation costs in GEMM and memory costs for SPMM and SDDMM. 

\vspace{.07in}
\textbf{Opportunity: Accelerating training by quantization.} Quantization offers two avenues to improve training speed, that is, higher computation throughput and less memory traffic with values in lower precision. We use quantized computing to accelerate the most computation-intensive primitives and operations, i.e., GEMM, SPMM and SDDMM. \textit{However, the problem is that these primitives are highly optimized and fine-tuned by commercial libraries. }And CUBLAS GEMM and cuSPARSE SPMM, and SDDMM are closed-source. Integrating our proposed optimizations into these kernels and achieving the desired speedup is extremely challenging. 


In this paper, (i) we utilize our novel quantization-aware GEMM to reduce computation time. Moreover, we identify an optimal tiling strategy to overlap the on-the-fly quantization of the matrix with the subsequent quantized computations. (ii) To address the memory-intensive nature of SPMM and SDDMM, we sequentially quantize the input tensor and write the quantized value in memory. The computation then randomly accesses the smaller quantized tensor, which provides better cache behavior than direct random access to full-precision tensors. 

\vspace{-.1in}
\subsection{Lightweight rules for maintaining training accuracy during quantized training}
\label{sec:acc}

\textbf{GPU-accelerated stochastic rounding.} We adopt stochastic rounding to reduce the quantization error~\cite{gupta2015deep}, with which the expectation of the quantization error should be 0 statistically. 
In particular, given a scaled floating-point number $x$ between $[-2^{B-1}-1, +2^{B-1}-1]$ as the range of $B$-bit integers, we round it to integer based on:
\begin{equation}
\small
    x_{Quant} = \left\{
    \begin{aligned}
          &floor(x) + 1, &w/ probability&\quad x - floor(x);  \\
          &floor(x), &w/ probability&\quad 1 - (x - floor(x)).
    \end{aligned}
    \right.
\normalsize
\end{equation}

We design and implement a GPU-accelerated pseudo-random number generator to facilitate fast stochastic rounding, which is $\sim$20$\times$ faster than the native cuRAND random number generator on GPU~\cite{cuRAND}. 
Our key optimization is storing random generator states in GPU registers as opposed to in global memory, which is the case in the existing cuRAND library~\cite{cuRAND}. Since the random number generator is a memory-bound operation, this optimization helps significantly improves the throughput. Of note, because cuRAND is closed-source, we cannot directly integrate this optimization into cuRAND. We thus implement our generator based upon xoshiro256++\cite{blackman2021scrambled} with our memory optimizations.


%

\textbf{Lightweight rule for deriving \# of desired quantization bits.}
We develop a metric to measure the quantization error, which subsequently helps derive the \# of desired quantization bits.
During quantization, a value $\textbf{X}_i$ will be rounded to one of the quantization grid points $\textbf{X}_{i,Quant}$.
We introduce the following metric to estimate the quantization error of a tensor \textbf{X}: 
\begin{equation}
    Error_{\textbf{X}} = \frac{1}{N}\sum^{N}_{i=1} \left|\frac{\textbf{X}_i - \textbf{X}_{i,Quant}}{\textbf{X}_i + \textbf{X}_{i,Quant}+\epsilon}\right|,
\label{eq:int}
\end{equation}
where $N$ is the number of elements in the tensor.

Intuitively, $Error_{\textbf{X}}$ derives the relative quantization error of a tensor \textbf{X}, where the numerator, i.e., $\left|\textbf{X}_i - \textbf{X}_{i,Quant}\right|$ is the absolute quantization error while the denominator is the sum of $\textbf{X}_i$, $\textbf{X}_{i,Quant}$, and $\epsilon$. The denominator needs the sum of the three values for two reasons: (i) a small $\epsilon$ to avoid dividing by zero, i.e., when $\textbf{X}_i=\textbf{X}_{i,Quant}=0$. {\gnn} chooses $\epsilon = 0.0005$. Of note, {\gnn} does not experience $\textbf{X}_i+\textbf{X}_{i,Quant}=0$ when $\textbf{X}_i\neq0$ and $\textbf{X}_{i,Quant}\neq0$ because our quantization is symmetric. (ii) If we use $\epsilon$ with only either $\textbf{X}_i$ or $\textbf{X}_{i,Quant}$ as the denominator, we could suffer from quantization error divided by $\epsilon$. This would lead to an extremely large relative error for a particular $\textbf{X}_i$, overshadowing the relative quantization error of other $\textbf{X}_i$'s. 

Our proposed quantization error metric in Equation~\ref{eq:int} is a relative error thus inductive. That is, this parameter could be used to compare the quantization error across tensors. Therefore, we can tune a desired $Error_{\textbf{X}}$ that is generally applicable for various tensors. 
The value range of $Error_{\textbf{X}}$ is [0, 1]. Particularly, if $\textbf{X}_i$ has no rounding error, the corresponding error is 0. When the rounding error of $\textbf{X}_i$ is significant, the term approaches 1.

We leverage Equation~\ref{eq:int} to select the desired number of quantization bits as follows:
we compute $Error_\textbf{X}$ of the output tensor of the first GNN layer with quantization. Note that we do not apply this metric to the input tensor of the first layer because its quantization error can be recovered by learning from the graph structure~\cite{abboud2020surprising}. 
We also want to mention that the training process could potentially amend the quantization error when the bit count is even lower. {Our bit count derivation metric derives a lower bound bit count that could maintain the training accuracy.}

\begin{figure}[ht]
\centering
\begin{tabular}{cc}  
    \subfloat[Training accuracy vs $Error_\textbf{X}$.]{
        \hspace{-.2in}
\includegraphics[width=.49\columnwidth]{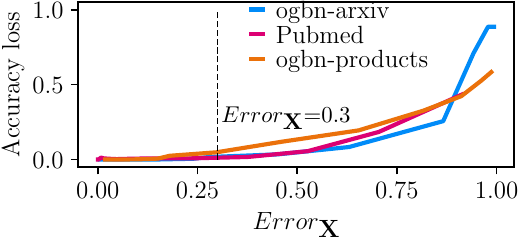}
\label{fig:tensormu}
    }
    &
    \subfloat[{$Error_\textbf{X}$ vs $B$}.]{
        \hspace{-.2in} 
        \includegraphics[width=.49\columnwidth]{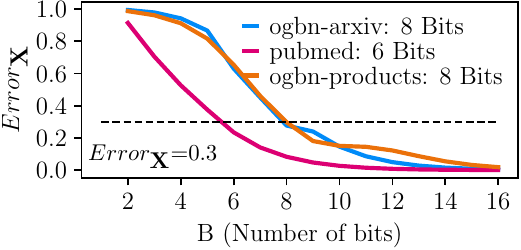}
        \label{fig:mubits}
    }
\end{tabular}
\caption{The accuracy for different $Error_\textbf{X}$ and the required number of bits to retain the desired $Error_\textbf{X}$ for ogbn-arxiv, Pubmed, and ogbn-products datasets. 
}
\vspace{-.05in}
\label{fig:beta}
\end{figure}

As shown in Figure~\ref{fig:tensormu}, our heuristic demonstrates that
when $Error_\textbf{X}$ < 0.3, {\gnn} can maintain the accuracy requirement across various datasets. Therefore, we let $Error_\textbf{X}$ = 0.3
across all datasets.
Figure~\ref{fig:mubits} shows that the desired number of bits for ogbn-arxiv, Pubmed, and ogbn-products are
8, 6, and 8, respectively.

The benefit of our metric is as follows: because our lightweight rule calculates the $Error_\textbf{X}$ solely for the first layer during the initial epoch. In contrast, determining the accuracy loss typically necessitates training the model until convergence (i.e., all epochs).
The effectiveness of our approach is demonstrated by our empirical findings presented in Fig~\ref{fig:beta}(a), which shows that $Error_\textbf{X}<=0.3$ is a general threshold to maintain the accuracy across datasets.

\textbf{Novel quantization-aware matrix multiplication with scaling factor computation.} Since the majority of the tensor primitives in GNN are either dense matrix multiplication or a variant of it, the accuracy analysis would be similar across these primitives. We hence restrict our accuracy analysis to dense matrix multiplication (i.e., GEMM) for brevity.



For two reasons, the resultant matrix of a quantized matrix multiplication has to be of higher precision. First, the result of a multiplication operation between two 8-bit integers could go beyond the value range of an 8-bit integer. Second, the subsequent accumulation of the multiplied values can again push the value beyond the range of an 8-bit integer. 


\begin{figure}[h]
    \centering
    \includegraphics[width=.35\textwidth]{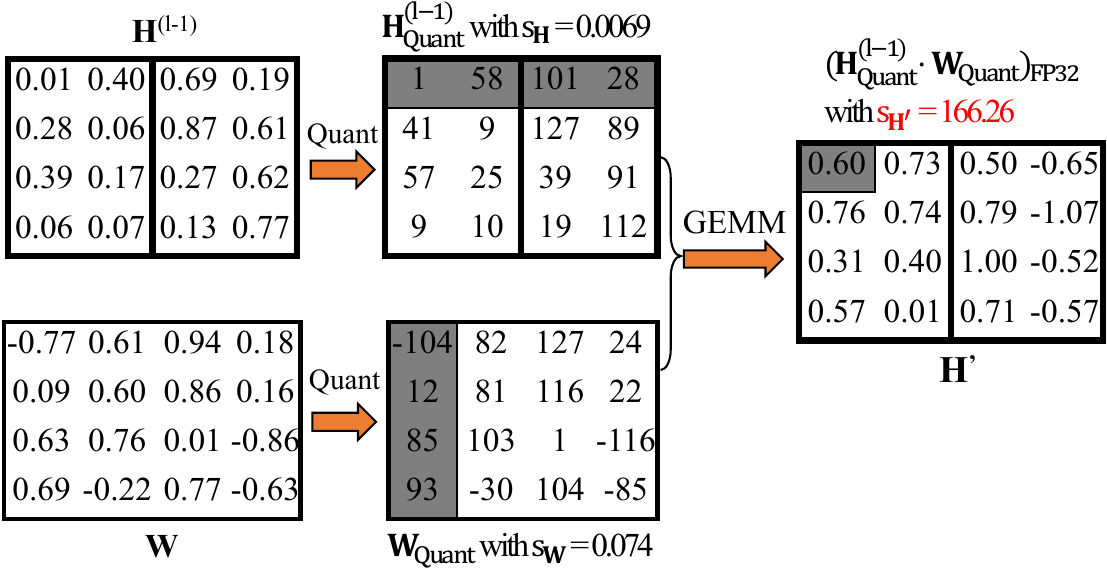}
    \caption{{Quantization for GEMM of step~\protect\circled{1} in Figure~\ref{fig:GAT_forward}}.
    }
    
    \label{fig:quant_method} 
\end{figure}

Figure~\ref{fig:quant_method} presents this problem when we perform $\textbf{H}^{(l-1)}\cdot \textbf{W}$ in quantized mode.
After quantization, the first row of $\textbf{H}^{(l-1)}_{Quant}$, i.e., [1 58 101 28] multiplies with the first column of $\textbf{W}_{Quant}$, i.e., [-104 12 85 93]$^T$ experiences both issues mentioned above. In fact, all the entries in the resultant matrix $(\textbf{H}^{(l-1)}_{Quant}\cdot \textbf{W}_{Quant})_{int32}$ exceed the 8-bit range of $[-127,127]$. 
Therefore, we opt for a 32-bit data format to store the result to avoid this overflow problem. The good news is that storing the results in 32-bit integers introduces negligible overheads on commodity GPUs. Also, note that recent tensor core units on NVIDIA GPUs force the resultant matrix to be a 32-bit integer matrix for the input of two 8-bit integer matrices. 

To reduce the quantization overheads, {\gnn} directly dequantizes the GEMM results, i.e., $\textbf{H}'$ into FP32 after computing the resultant matrix with our optimizations. In the meantime, {\gnn} also derives the scaling factor $s_{\textbf{H}'}$=166.26 during the quantized GEMM operation, as shown in Figure~\ref{fig:quant_method}. This design avoids a dedicated dequantization kernel, a scaling factor computation kernel, and the associated expensive global memory accesses.

\textbf{Full precision weight update.} To combat the round-off error, we update the model weights with dequantized FP32 gradients. The reason is that 
the magnitude of the model weight is often significantly larger than the gradients. In addition, the small learning rate further amplifies the difference. 
Previously, existing projects use shared exponent~\cite{sakr2018per}, Flexpoint~~\cite{koster2017flexpoint}, or delayed updates~\cite{park2018training} to tackle the round-off error. Unfortunately, these designs could suffer from shortcomings of delayed convergence, unavailability on commodity GPUs, being slow to implement on GPUs, or multiple of these shortcomings~\cite{chen2017fxpnet,zhou2016dorefa}. Of note, although the updated FP32 weights are quantized into 8-bit integers in the next iteration, quantizing the updated weights into 8-bit integers is often better than directly updating the quantized weights with quantized gradients {as elaborated below.}

Assuming $W_{full}=W_{quant}+W_{round\_off}$ and $\Delta W_{full}=\Delta W_{quant}+\Delta W_{round\_off}$, where $W_{full}$ and $\Delta W_{full}$ are weights and update values (e.g. gradients) of full precision, respectively. 
$W_{quant}$ and $\Delta W_{quant}$ are the output from the quantization function $Q$.
$W_{round\_off}$ and $\Delta W_{round\_off}$ are the correspondingly round-off errors.
Below is our analysis: 
\begin{equation}
    Q(W_{full}) + Q(\Delta W_{full}) = W_{quant} + \Delta W_{quant}.
\label{eq:quant_add}
\end{equation}
If we add the full precision before quantization, we arrive at: 
\begin{equation}
\begin{split}
    Q(W_{full}+\Delta W_{full}) &\sim W_{quant} + \Delta W_{quant}\\
    &+Q(W_{round\_off}+\Delta W_{round\_off}).
\end{split}
\label{eq:add_quant}
\end{equation}
One can observe that Equation~\ref{eq:add_quant} offers higher accuracy than Equation~\ref{eq:quant_add} as $Q(W_{round\_off}+\Delta W_{round\_off})$ curbs the round off error. 


\textbf{Full precision for the layer before Softmax.}
The Softmax layer amplifies the quantization error due to its exponential operations.
For simplicity, we consider a layer before Softmax with two outputs $z_0$ and $z_1$. The difference in Softmax score ($D$) between the two outputs is:  
\begin{equation}
    D = \frac{\frac{exp(z_0)}{exp(z_0)+exp(z_1)}}{\frac{exp(z_1)}{exp(z_0)+exp(z_1)}}=\frac{exp(z_0)}{exp(z_1)}.
\end{equation}
Once the quantization error $e_i$ is introduced to $z_i$, the perturbation of output difference follows: 
\begin{equation}
    D' = \frac{exp(z_0+e_0)}{exp(z_1+e_1)}=D\cdot\frac{exp(e_0)}{exp(e_1)}=D\cdot \underbrace{exp(e_0 - e_1)}_{\textrm{Amplified\ error}}.
\end{equation}
This analysis suggests that the exponential function applied to $(e_0-e_1)$ will rapidly make the difference $D$ either bigger or smaller, departing from the desired faithful $D$.
Therefore, we propose to use full precision to compute the layer before the Softmax.

\subsection{Quantization accelerated training} 
\label{subsec:fast}

\begin{figure}[h]
    \centering
    \hspace{-.1in}\includegraphics[width=.45\textwidth]{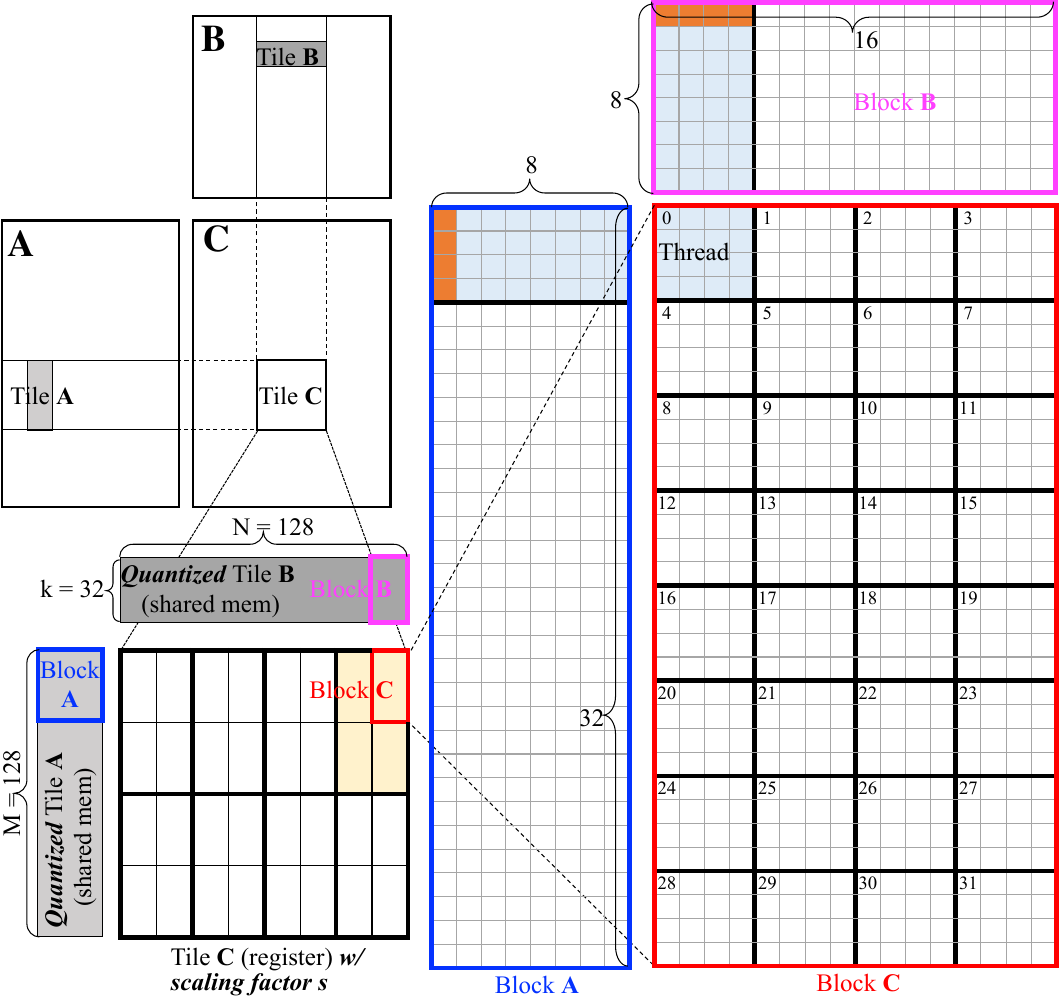}
    \caption{{\gnn} GEMM with on-the-fly quantization and scaling factor s computation.  
    }
    \label{fig:QGEMM}
\end{figure}

\textbf{GEMM with on-the-fly quantization.}
Figure~\ref{fig:QGEMM} illustrates our GEMM with on-the-fly quantization and scaling factor (i.e., s) computation. Our quantized GEMM includes four steps:
First, we quantize while loading the tiles of input matrices from global memory to shared memory (i.e., Tiles \textbf{A} and \textbf{B}). Note that the input matrices are usually needed for backward computation. Therefore, we store the quantized tiles back in global memory while computing, eliminating the round-trip memory latency in the naive design. Second, we store the resultant block in registers to minimize the write latency. Third, during computation, we pack four 8-bit integers into a 32-bit register and use one \textbf{DP4A} instruction for four multiply-accumulate operations between two packed registers. Fourth, we dequantize the resultant 32-bit integers in registers to 
floating-point. We also fuse the computation of parameter $s$ in the kernel for the following primitives.

We develop a data tiling strategy to hide the data access latency behind the computations. First, when loading from global memory, we choose an appropriate tile size, i.e., $128\times32$ which is the ideal tile size to balance the computation capability and memory throughput on V100 GPUs. 
Below is our analysis: assuming the sizes of Tiles \textbf{A} and \textbf{B} are $M\cdot k$ and $N\cdot k$, to pipeline the loading of Tiles \textbf{A} and \textbf{B} with the computation, we hide the loading latency by arithmetic operations, $loadLatency=arithmeticLatency$.
We denote that the latency of loading one FP32 value from global memory is $Latency_{global}$ and the latency of performing one multiply-accumulate operation is $Latency_{compute}$. We arrive at $loadLatency=(M\cdot k + N\cdot k) \cdot Latency_{global}$, and $arithmeticLatency=(M\cdot N\cdot k) \cdot Latency_{compute}$.
This leads to $\frac{M+N}{M\cdot N}=\frac{Latency_{compute}}{Latency_{global}}$. 
On V100 GPU, we find $Latency_{global}\approx 400$, $Latency_{compute}\approx 4$. Without loss of generality, we let $M=N$. We hence arrive at $M=N\approx200$. Since the size of $M$ and $N$ should be the power of two, we find $M=N=128$ offer the best performance. 
Second, at the warp level, we let each warp load two blocks from Tiles \textbf{A} and \textbf{B} to compute four adjacent blocks in Tile \textbf{C} as shown in Figure~\ref{fig:QGEMM}. We derive the optimal block size that can hide the latency of accessing shared memory. Particularly, in each iteration, a thread loads 32 packed INT8 as the eight blocks colored on the right side of Figure~\ref{fig:QGEMM}. Then the 16 \textbf{DP4A} instructions cover the 18 cycles latency of loading for the next iteration.

For computation, we carefully schedule the threads to improve the computation intensity. First, when loading and quantizing Tile \textbf{A} from global memory to shared memory, we transpose the Tile because the access is in column while Tile \textbf{A} is row-major in global memory. Second, to avoid bank conflict, a warp stores Block \textbf{A} in shared memory with a 16-byte offset in each column. Each warp works on $2\times 2$ \textbf{C} blocks to reuse Block \textbf{A} and \textbf{B}. Third, we schedule the threads as shown in the left side of Figure~\ref{fig:QGEMM}, mapping 32 threads within a warp to block \textbf{C} to increase shared memory throughput. For example, threads 0,1,2,3 access the same address in block \textbf{A} so the loaded data can be broadcast to 4 threads.

Of note, there exist frameworks that can generate GEMM kernels with efficient tiling and scheduling, but none of them can be used to implement GEMM with \textit{on-the-fly quantization}. On the one hand, template-based frameworks, such as AutoTVM~\cite{chen2018learning} and Ansor~\cite{zheng2020ansor}, optimize kernels by enumerating  the combinatorial choices of optimizations (e.g., tile layout, tile size, and parallelization). Searching the design space is time-consuming, and the generated kernels are not guaranteed to be optimal. On the other hand, on-the-fly quantization is not supported by the existing templates. Moreover, the kernel generated by these frameworks, e.g., triton compiler~\cite{tillet2019triton}, uses too many registers, resulting in unsatisfied performance.

\textbf{Incidence matrix-based adaptive SPMM.}
{\gnn} performs quantization in a separate kernel for SPMM.
We use quantization to reduce the memory traffic for SPMM since quantization leads the node and edge feature matrices to a smaller size. 
Unlike GEMM, which performs sequential memory access for the input matrices, SPMM experiences random memory accesses. 
In this case, performing on-the-fly quantization would lead to random memory access for input matrices of 32-bit floating-point data type. Further, because of unpredictable access patterns, on-the-fly quantization could potentially lead to repeated quantization of the same data. Instead, a dedicated quantization kernel would read 32-bit input floating-point matrices sequentially once and write the 8-bit quantized matrices out, again, sequentially and once. Therefore, during SPMM, we perform random memory access to input matrices of 8-bit as opposed to 32-bit. 



{\gnn} further introduces two SPMM variant optimizations, that is, incidence matrix-based SPMM and adaptive SPMM to improve the quantized training performance. 


 \begin{figure}[h]
\centering
        
\begin{tabular}{c}  
    \subfloat[DGL's step \protect\circled{3''} of Figure~\ref{fig:GAT_back}.]{
        
\includegraphics[width=.75\columnwidth]{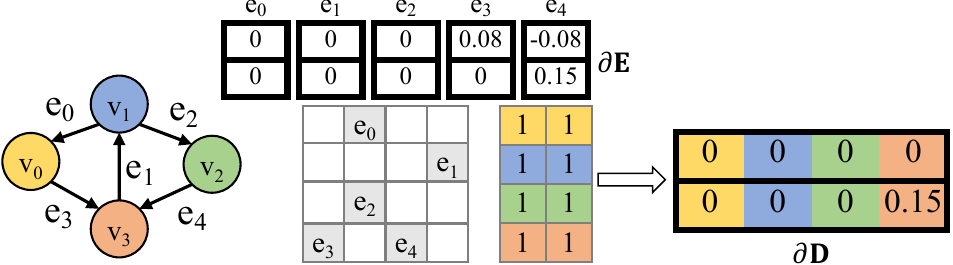}
\label{fig:dgl_incidence}
    }\\
    \subfloat[{\gnn}'s incidence matrix-based step \protect\circled{3''} of Figure~\ref{fig:GAT_back}.]{
        \includegraphics[width=.8\columnwidth]{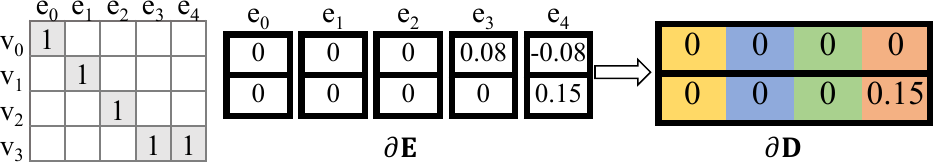}
        \label{fig:our_incidence}
    }
\end{tabular}
\caption{Incidence matrix-based SPMM.}
\label{fig:incidence:all}
\end{figure}


\textbf{Incidence matrix-based SPMM} accelerates the SPMM variant, i.e., step \circled{3''} of Figure~\ref{fig:GAT_back}. Particularly, this SPMM computes the gradients of node features by aggregating the incoming edge features for each node. Using node $v_3$ as an example, as shown in Figure~\ref{fig:dgl_incidence}, because $v_3$ contributes to the edge features for $e_3$ and $e_4$, its gradient is the partial derivative of $e_3$ and $e_4$, that is, $\partial {v}_3 = \partial e_3 + \partial e_4$.
However, because DGL uses the adjacency matrix format for the graph, shown in Figure~\ref{fig:dgl_incidence}, we need three matrices for the SPMM, that is, this graph, $\partial e$, and the node features with all ``1''s. 

%

The drawback of this design is two-fold: First, one needs to allocate and access the all ``1'' node feature matrix which is redundant and expensive.
Second, although the operation in Figure~\ref{fig:dgl_incidence} can be formulated as an SPMM operation, it includes three inputs, which is not supported by the state-of-the-art cuSPARSE library.

In Figure~\ref{fig:our_incidence}, {\gnn} formulates this \circled{3''} computation as an incidence matrix-based SPMM. 
Particularly, the incidence matrix is a $V\times E$ matrix where $V$ and $E$ are, respectively, the number of nodes and edges in the graph. Each row of the incidence matrix contains the incoming edges of each node by marking the associated entry as 1. This design allows us to compute step \circled{3''} by multiplying two matrices, i.e., the incidence matrix and edge feature. Because we only need two input matrices, {\gnn} can now adopt high-performance cuSPARSE SPMM kernels for step \circled{3''}, which is significantly faster than DGL's three matrices-based SPMM.

\begin{figure}[h]
    \centering
    \includegraphics[width=.45\textwidth]{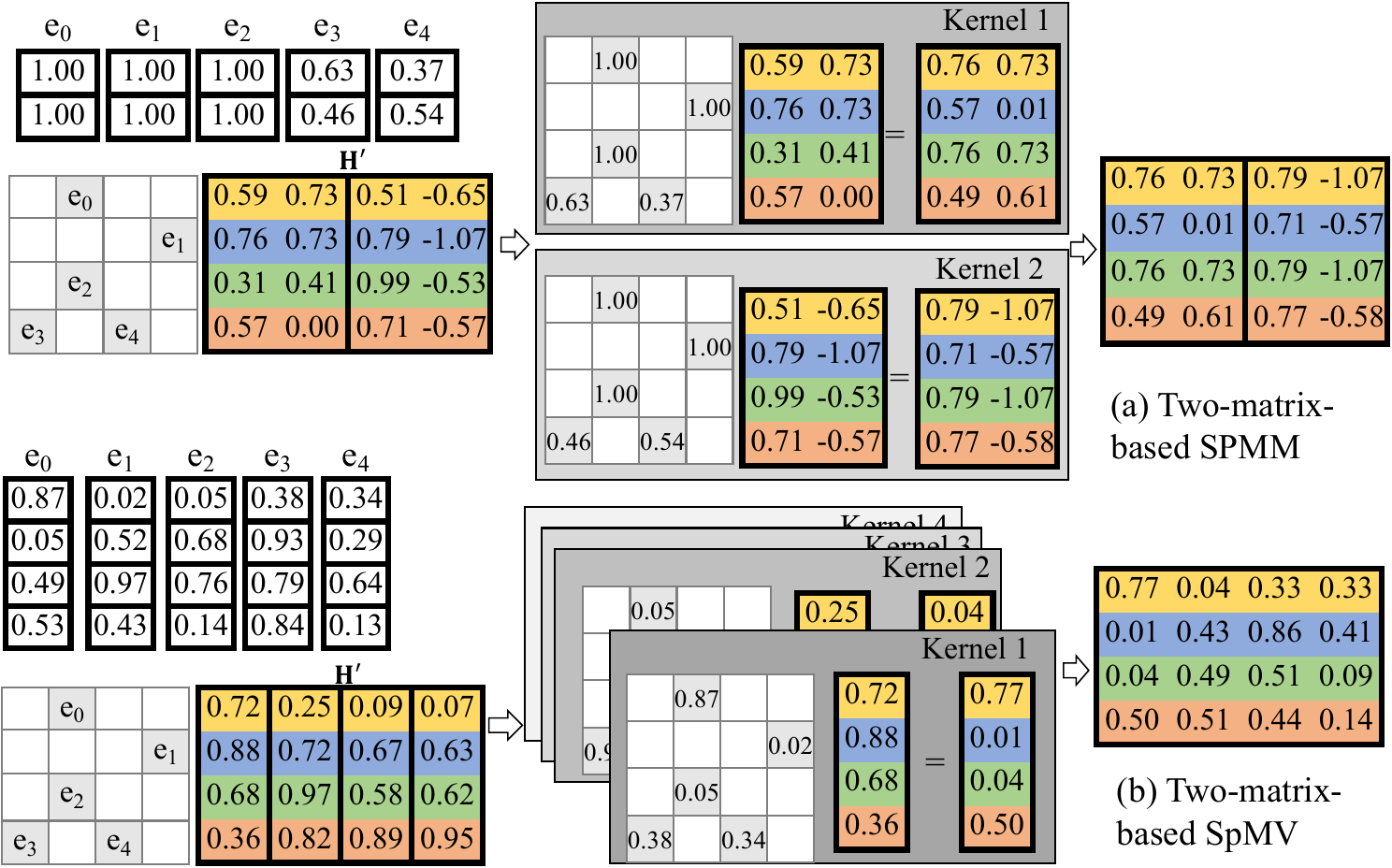}
    \caption{Transforming three-matrix-based SPMM, e.g., step \protect\circled{5} in Figure~\ref{fig:GAT_forward} into a collection of: (a) two-matrix-based SPMM and (b) two-matrix-based SpMV.}
    
    \label{fig:SPMM} 
\end{figure}

\textbf{Kernel count-based adaptation.}
Chances are certain SPMM computations still involve three matrices, such as step \circled{5} in Figure~\ref{fig:GAT_forward}. 
In Figure~\ref{fig:SPMM}, we demonstrate how one could transform a three-matrix-based SPMM kernel
into a collection of two-matrix-based SPMM kernels or, to an extreme, Sparse Matrix-Vector multiplication (SpMV) kernels. Once that transformation is completed, one could directly rely on cuSPARSE to perform each two-matrix-based SPMM or SpMV. Note, we prefer cuSPARSE over DGL primitives because our evaluation shows that a single cuSPARSE SPMM kernel is significantly faster than DGL's native two-matrix-based SPMM across various configurations.

However, the benefits brought by cuSPARSE kernels diminish along with the increment of the number of kernels, as kernel invocation cost soars when too many kernels are launched. In summary, neither DGL nor transformed cuSPARSE bests the other across all configurations. We hence adaptively leverage these two solutions to achieve the best performance of both worlds. 

Figure~\ref{fig:SPMM}a depicts the SPMM with both edge and node features as matrices. 
In this design, the first head of the node feature is scaled by the first element of the edge feature and similarly for the second head. We can use multiple optimized cuSPARSE SPMM kernels to replace the native kernel used in DGL. 
In this case, the two heads can be finished by two SPMM kernels. 
%
Figure~\ref{fig:SPMM}b  assumes we have four heads in step \circled{5} of Figure~\ref{fig:SPMM}. In this case, we arrive at four SpMV kernels for the original three-matrix-based SPMM.

\textbf{SDDMM with on-the-fly dequantization.} Similar to our SPMM design, we first perform sequential memory access to quantize the input matrices. During SDDMM, we perform random memory access on those quantized matrices of smaller sizes. Briefly, the existing SDDMM performs one round of random access on full precision matrices. In contrast, {\gnn} performs one round of sequential access on full precision matrices and one round of random access to the low precision matrices. This will lead {\gnn} to have a shorter turnaround time. 
Since SDDMM might perform addition or subtraction operations, one cannot directly compute the quantized values. This leads to our SDDMM with on-the-fly dequantization.


We use step \circled{3} of Figure~\ref{fig:GAT_forward} to explain the reason. This SDDMM computes the edge feature by $\textbf{E}[i][j] = \mathbf{S}[v_i] + \mathbf{D}[v_j]$. 
Assuming the scaling factor $s$ for \textbf{S} and \textbf{D} are, respectively, $s_{\textbf{S}}$ and $s_{\textbf{D}}$.
The addition in quantized format should be $\mathbf{{S}}[v_i] + \mathbf{{D}}[v_j] \approx s_{\textbf{S}}\cdot\mathbf{S}_{Quant}[v_i] + s_{\textbf{D}}\cdot\mathbf{D}_{Quant}[v_j]$. Because $s_{\textbf{S}}$ and $s_{\textbf{D}}$ are often not equal, one cannot directly add the quantized values $\mathbf{S}_{Quant}[v_i]$ and $\mathbf{D}_{Quant}[v_j]$.
Therefore, {\gnn} loads the quantized data to enjoy reduced memory traffic, subsequently on-the-fly dequantize the loaded values for addition/subtraction computation. 

If SDDMM performs multiplication and division, we can conduct SDDMM directly on the quantized value. Using step \circled{5''} of Figure~\ref{fig:GAT_back} as an example, one needs to compute $\partial\boldsymbol\alpha[e_0]=\partial \mathbf{H}^{(l)}[v_0]\cdot \mathbf{H}'[v_1]$. In this case, assuming the scaling factor of $\partial \mathbf{H}^{(l)}[v_0]$ and $\mathbf{H}'[v_1]$ are $s_0$ and $s_1$. The computation can be approximated as 
${\partial\boldsymbol\alpha}[e_0]\approx (s_0\cdot\partial\mathbf{H}^{(l)}[v_0])\cdot (s_1\cdot \mathbf{H}'[v_1])$. We can further arrive at $\partial\boldsymbol\alpha[e_0]\approx (s_{0}\cdot s_{1})\cdot\partial \mathbf{H}^{(l)}[v_0]\cdot \mathbf{H}'[v_1]$. This allows {\gnn} to perform quantized multiplication directly. Division can also directly work on the quantized values.  

\textbf{Inter-primitive optimization}. Noticing that the follow-up operators can reuse some quantized tensors, {\gnn} caches these quantized tensors to reduce the quantization overhead. In general, there exist two caching opportunities: (1) caching forward pass for backward, (2) caching prior operators for subsequent operators. {\gnn} develops a detection algorithm that runs on the computation graphs to automatically derive these reuse cases. 

First, the backward computation can reuse the quantized tensors from the forward pass. 
For example, the GEMM {of step \circled{1} in Figure~\ref{fig:GAT}} has the forward computation $\textbf{H}' = \textbf{H}^{(l-1)} \cdot \textbf{W}$. The corresponding backward step contains the gradient computation for weight, that is, $\partial \textbf{W} = \textbf{H}^{(l-1)} \cdot\partial\textbf{H}'^T$, and the gradient computation of node features, i.e., $\partial \textbf{H}^{(l-1)} = \partial\textbf{H}' \cdot \textbf{W}^T$, as shown in step \circled{1'} in Figure~\ref{fig:GAT_back}. Clearly, the quantized matrices $\textbf{H}^{(l-1)}$ and \textbf{W} are used in both forward and backward computations. We {thus} save the quantized input $\textbf{H}^{(l-1)}$ and \textbf{W}  during the forward pass for the backward pass to avoid repeated quantization. Second, when two operators share the same tensor as input, we can cache the quantized tensor from the former operator and use it for the latter. For example, in Figure~\ref{fig:GAT_back}, the SPMM in step \circled{5'} and SDDMM in \circled{5''} both need the quantized $\partial \mathbf{H}^{(l)}$. This way, we cache the quantized input tensor. We also intentionally schedule the computation orders such that  the cached tensors can be reused.

We derive the caching opportunity on the computation graph, i.e., Figure~\ref{fig:GAT_forward} as follows. The computation graph consists of tensors as nodes and operators as edges. For nodes with more than one out edge, we can quantize once for multiple operators. For example, the tensor $\partial \textbf{H}^{(l)}_{\text{Quant}}$ in Figure~\ref{fig:GAT_back} has two operators as out-edges, so we cache this tensor. Then we reverse the edges in the computation graph for the backward pass. In this backpropagation graph, we will check if the to-be-quantized tensors are already quantized in the forward graph in order to facilitate quantization sharing.

\textbf{Quantization overhead vs. benefit analysis.}
While quantization helps reduce computation and data movement, it also brings overheads. Mainly, quantization introduces two types of overheads: parameter computing and data type casting. 
First, the parameter $s$ in Equation~\ref{eq:back:quant} is derived by reducing the elements with the maximum absolute value. For a $N\times N$ matrix, the reduction needs $N^2$ operations to derive the absolute maximum values. Second, quantizing or dequantizing an element requires two operations: multiply and data type casting. Therefore, the quantization before the primitive performs $4N^2$ floating-point operations.

%

For GEMM with the input matrices at sizes of $M \times K$ and $K \times N$, we perform $4K(M+N)$ and $2MN$ operations for quantization and dequantization, respectively. 
For GEMM, we reduce the number of multiply-accumulate instructions from $MNK$ to $\frac{MNK}{4}$, which is often significantly higher than the overheads. 
Sparse primitives quantize the node and edge feature matrices. We assume $D$ as the size of features. Given a graph with $N$ nodes and $E$ edges, in SPMM, the node and edge features require $4D(N + E)$ operations for quantization. Later, only the resultant node features are dequantized with $2ND$ operations. Quantization in SDDMM performs $4ND$ operations for node features. Dequantizing resultant edge features needs $2ED$ operations.
Regarding the benefits of sparse primitives, sparse primitives enjoy a better cache access pattern brought by quantization which reduces the sizes of the input matrices. 


\begin{figure*}[ht]
\begin{tabular}{ccccc}  
    \subfloat[Pubmed GCN.]{
        \hspace{-.2in}
        \includegraphics[width=.19\linewidth]{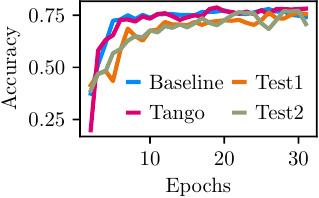}
    }    
    &
    \subfloat[ogbn-arxiv GCN.]{
        \hspace{-.2in} 
        \includegraphics[width=.19\linewidth]{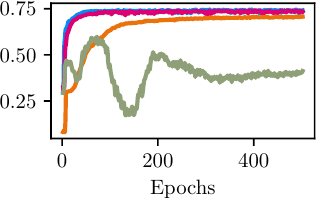}
    } 
    &
    \subfloat[\hspace{-.01in}ogbn-products GCN.]{
        \hspace{-.2in}
        \includegraphics[width=.19\linewidth] {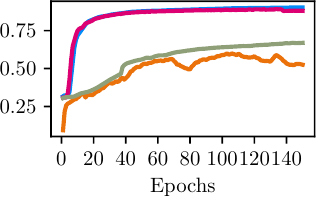}
    }
    &
    \subfloat[DBLP GCN.]{
        \hspace{-.2in}
        \includegraphics[width=.19\linewidth]{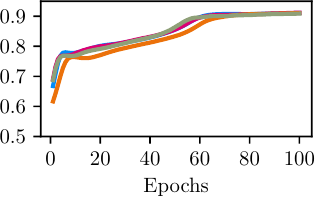}
    }  
    &
    \subfloat[Amazon GCN.]{
        \hspace{-.2in}
        \includegraphics[width=.19\linewidth]{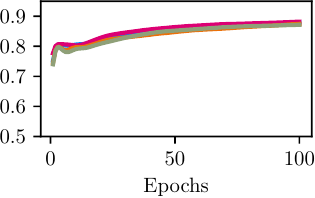}
    } 
    \\
    \subfloat[Pubmed GAT.]{
    \hspace{-.2in}
        \includegraphics[width=.19\linewidth]{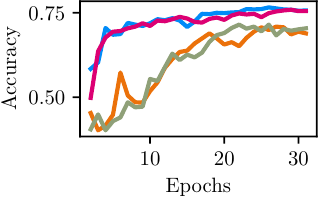}
    }   
    &
    \subfloat[ogbn-arxiv GAT.]{
        \hspace{-.2in}
        \includegraphics[width=.19\linewidth]{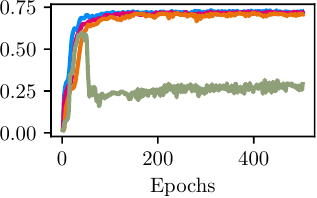}
    } 
    &
    \subfloat[\hspace{-.01in}ogbn-products GAT.]{
        \hspace{-.2in}
        \includegraphics[width=.19\linewidth]{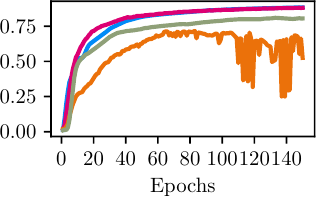}
    }  
    &
    \subfloat[DBLP GAT.]{
        \hspace{-.2in}
        \includegraphics[width=.19\linewidth]{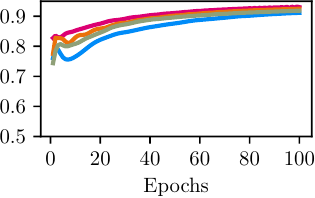}
    }  
    &
    \subfloat[Amazon GAT.]{
        \hspace{-.2in}
        \includegraphics[width=.19\linewidth]{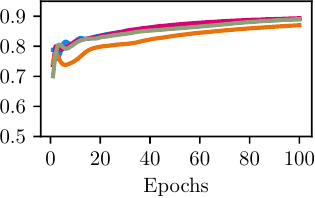}
    }  
\end{tabular}
\caption{{The convergence analysis of GCN and GAT with {\gnn}. \textbf{Test1} denotes {\gnn} with quantized layer before Softmax. \textbf{Test2} denotes {\gnn} using nearest rounding instead of stochastic rounding}. }

\label{fig:eva:quant}
\end{figure*}

\section{Experiments}\label{sec:eval}
%

\subsection{Experimental setup}
%

\begin{table}[ht]
\centering
\resizebox{\columnwidth}{!}{%
\begin{tabular}{c|c|c|c|c|c}
\hline
Dataset       & ogbn-arxiv & ogbn-products & Pubmed       & DBLP      & Amazon \\ \hline
Nodes         & 169,343    & 2,449,029 & 19,717     & 317,080   & 410,236                    \\ \hline
Edges         & 1,166,243  & 61,859,140 & 88,651  & 1,049,866 & 3,356,824                  \\ \hline
{Task} & {NC}     & {NC}  & {NC}        & {LP}       & {LP}                        \\ \hline
\end{tabular}%
}
\caption{{Graph datasets. NC denotes ``node classification'', and LP denotes ``link prediction''.}}
\vspace{-.3in}
\label{tb:dataset}
\end{table}

\textbf{Datasets.} We conduct experiments on five graph datasets as shown in Table~\ref{tb:dataset}. The \textit{obgn-arxiv}~\cite{wang2020microsoft} and \textit{Pubmed}~\cite{namata2012query} are citation graphs whose nodes represent papers and edges are the citations. The task is to predict the categories of papers. 
The \textit{ogbn-products}~\cite{Bhatia16} dataset depicts an product co-purchasing network, where nodes represent products sold in Amazon, and edges between two products indicate that they are purchased together. The task is to predict the category of a product.
The \textit{DBLP} dataset is a co-authorship network where nodes are authors and edges are co-authorship~\cite{yang2015defining}. The \textit{Amazon} dataset contains products as the nodes and edges represent co-purchase~\cite{leskovec2007dynamics}. 
The tasks of \textit{DBLP} and \textit{Amazon} are to predict if a link exists between two nodes.
We add the reverse edges for the directed graphs and self-connects edges to ensure the SPMM operation works for every node. 

%



\textbf{Models.}
We evaluate GCN~\cite{kipf2016semi} and GAT~\cite{GAT2018graph} from the example implementations of DGL and the models are trained with the same number of epochs and hyperparameter settings. Both models use the hidden size of 128 and two GNN layers; GAT has four attention heads. 
For node classification, the model generates the node embedding as the set probabilities of each category. For link prediction, we perform dot-product between two node embeddings as the score of edge existence. 
The training epochs for  Pubmed, ogbn-arxiv, and ogbn-products are 30, 500, and 150. 




\textbf{Implementation details}. \textit{For ease of use, we integrate {\gnn} in DGL. Therefore, all the models on DGL can enjoy the performance benefits brought from {\gnn} without any changes.} We provide our optimized quantized CUDA kernels to replace the corresponding primitives in DGL. 
DGL employs the GEMM function from the cuBLAS library and sources its sparse primitives either directly from cuSPARSE~\cite{cusparse} or through its own implementations. DGL's interface is designed in Python and its primitives are integrated as PyTorch functions. To ensure a fair comparison, the kernels of {\gnn} are also invoked via PyTorch's auto-differential engine during training~\cite{pytorch}. Furthermore, DGL supports multiple graph data structure formats, and {\gnn} leverages DGL's heuristics to determine the most efficient format for its primitives.





\textbf{Evaluation platforms.} We use Python 3.6.10 and CUDA 11.7 on six V100S GPUs and Intel(R) Xeon(R) Gold 6244 @ 3.60GHz CPU. The model is trained with PyTorch 1.13.0 and DGL 0.8. We also have access to a single A100 GPU for a limited time. We use that to compare GEMM on INT8 tensor core vs FP16 tensor core.

%

%

%
%

%
%

%
\subsection{{\gnn} vs. state-of-the-art}
We compare {\gnn} with DGL~\cite{wang2019dgl} and EXACT~\cite{liu2022exact}. DGL trains the model in full precision, and EXACT trains the model with quantized tensors. EXACT aims to save memory by quantizing the saved tensors, and it has no optimization in computation. We set EXACT to use 8-bit quantization. Of note, the GNN models are implemented through PyTorch, which experiences significant high-level language overheads when calling {\gnn} primitives. As a result, we observe significantly smaller model-level speedups than the primitive-level comparisons (detailed in Section~\ref{subsec:eval:turnaround}).

\begin{figure}[h]
\centering
\includegraphics[width=.9\columnwidth]{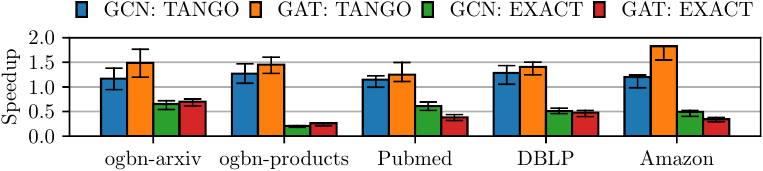}
\caption{The speedup of training the GNN models with {\gnn} and EXACT compared with DGL.}
\vspace{-.1in}
\label{fig:eva:model}
\end{figure}

\textbf{Training speed.} We evaluate the training speedup of {\gnn} on GCN and GAT models. We train each model 5 times and report the average elapsed time achieving the same accuracy as the baseline, including the forward and backward computations. As shown in Figure~\ref{fig:eva:model}, {\gnn} has $1.2\times$ and $1.5\times$ speedup on average on GCN and GAT models compared with DGL, respectively. 
The GAT model enjoys more benefits because it contains more quantized primitives than GCN. 
Further, larger graphs have more speedup for the GCN model, except the DBLP dataset, which has the smallest average degree among the five graphs. Overall, {{\gnn} achieves an average speedup of $2.9\times$ on GCN and $4.1\times$ on GAT compared with EXACT. The key takeaway is that applying quantization without appropriate optimizations will lead to a significant slowdown (e.g., EXACT).}

\textbf{Accuracy study.} Figure~\ref{fig:eva:quant} studies the accuracy impact of the techniques in Section~\ref{sec:acc} for GCN and GAT. In particular, we evaluate {\gnn}, {\gnn} with quantized layer before Softmax (\textbf{Test1}), and {\gnn} without stochastic rounding (\textbf{Test2}). For clarification, the training crashes without quantization-aware matrix multiplication and full precision weight update.
The baseline models are trained in FP32 with the same number of epochs as the quantized training. 

Overall, 
{\gnn} could achieve >99\% accuracy of the full precision training with the same number of epochs. 
When quantizing the layer before Softmax (\textbf{Test1}), the models show noticeable accuracy loss except for the DBLP dataset, GCN model on Pubmed, and GAT model on ogbn-arxiv. The average relative accuracy drop is 9.7\%. 
Despite the similar final accuracy, GAT on Pubmed and GCN on ogbn-arxiv converge slower than the baseline by 18 and 35 epochs, respectively.
For quantization without stochastic rounding (\textbf{Test2}), we observe for GCN on Pubmed and both models on DBLP and Amazon, the quantization error changes the optimization direction in some epochs. Thus the models take more epochs to recover. Moreover, the models on ogbn-arxiv and ogbn-products suffer from significant accuracy drops. As shown in Figure~\ref{fig:eva:quant}a, although the model can achieve convergence, the training process shows more instability than that with stochastic rounding. 

\begin{figure}[h]
    \centering

\includegraphics[width=0.95\columnwidth]{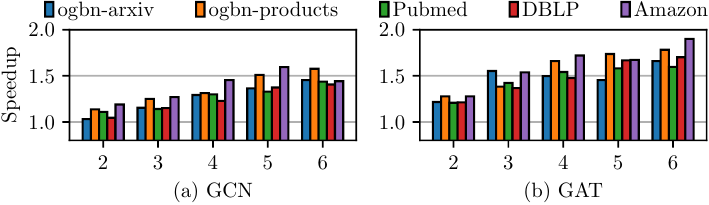}
\caption{
{\gnn}'s impact on multi-GPU training. The X-axis is the number of GPUs.}

\label{fig:multiGPU}
\end{figure}

%

\textbf{Multi-GPU training.} Figure~\ref{fig:multiGPU} studies {\gnn}'s impact on multi-GPU training. We directly adopt DGL's mini-batch multi-GPU training. 
That is, each GPU trains the model on a batch of sampled sub-graphs per epoch. Then, the gradients of all GPUs are updated by an all-reduce operation. \textit{We compare the training speed between full precision baseline and {\gnn} using the same number of GPUs.}

{\gnn} achieves speedup over the full precision baseline via transferring the quantized node features and gradients. Since we perform stochastic rounding-based quantization, this process will introduce nontrivial turnaround time. In {\gnn}, we overlap the feature quantization with the subgraph sampling. The overall trend is that more GPUs would enjoy higher speedup as the Peripheral Component Interconnect Express (PCI-E) congestion is better alleviated by our quantization. Particularly, the speedup increases from $1.1\times$ to $1.5\times$, and $1.2\times$ to $1.7\times$ from two to six GPUs on GCN and GAT, respectively. 



\subsection{Turnaround time analysis}\label{subsec:eval:turnaround}

%
%

\begin{figure}[h]
\begin{center}
\includegraphics[width=.9\columnwidth]{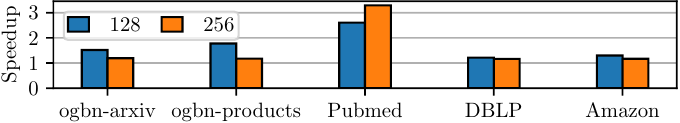}
\end{center}
\caption{The speedup of caching the quantized tensors.
}
\vspace{-.1in}
\label{fig:eva:cache}
\end{figure}

\textbf{Caching the quantized tensors.}
Figure~\ref{fig:eva:cache} shows the performance of caching the quantized tensor in forward for backward reuse. We test the GEMM primitive on different datasets. We test with two hidden sizes, $D = 128$ and $D = 256$. The result shows $1.7\times$ and $1.6\times$ on average when $D = 128$ and $D = 256$, respectively. The saving is related to the data size; smaller graphs, such as Pubmed, enjoy more time savings. 

\begin{figure}[h]
\centering

\subfloat[CUDA core (V100S).]{
    \includegraphics[width=.9\columnwidth]{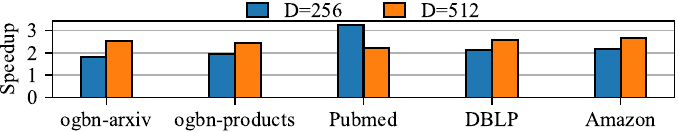}
    \label{fig:GEMM_a}
}

\subfloat[Tensor core (A100).]{
    \includegraphics[width=.9\columnwidth]{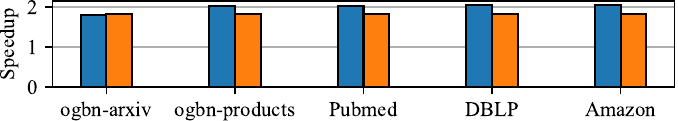}
    \label{fig:GEMM_b}
}

\caption{{\gnn} GEMM vs cuBLAS GEMM.}
\vspace{-.1in}
\label{fig:eva:GEMM}
\end{figure}

\textbf{GEMM.}
Figure~\ref{fig:GEMM_a} shows the speedup of our quantized GEMM over cuBLAS GEMM with the hidden size $D=256$ and $D=512$. {Of note, we include the quantization cost in {\gnn} GEMM time.}
Our quantized GEMM primitive has $2.2\times$ and $2.5\times$ speedup on average when $D=256$ and $D=512$, respectively. 
The trend also suggests that quantization offers more speedup on the GEMM operator when the hidden size increases. 
In addition, we also compare our quantized INT8 GEMM with FP16 GEMM on A100 Tensor Core GPUs. Figure~\ref{fig:GEMM_b} shows that our quantized GEMM primitives have $1.9\times$ for $D=256$ and $1.8\times$ for $D=512$.
Since both baseline and {\gnn} use Tensor Core, the speedup over baseline is smaller than using CUDA core because the performance difference of computing in INT8 and FP16 is 2$\times$ on A100 tensor cores.
Further, we observe a speedup drop for a bigger $D$ because our quantization needs to scan through a bigger tensor to extract the scaling factor $s$. 

\begin{figure}[h]
\centering
    \subfloat[{Throughput.}]{
        \includegraphics[width=.9\columnwidth]{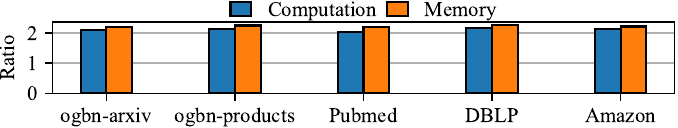}
        \label{fig:GEMM:thru}
    }
    
    \subfloat[{Instruction.}]{
        \includegraphics[width=.9\columnwidth]{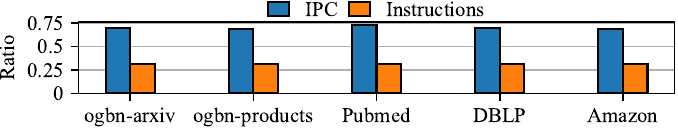}
        \label{fig:GEMM:inst}
    }
      \vspace{-.1in}
   \caption{The hardware profiling of quantized GEMM.}
    \label{fig:eva:GEMM_profile}
\end{figure}

{Figure~\ref{fig:eva:GEMM_profile} shows the profiling results of quantized GEMM. We profile the ratio of achieved computation throughput (operation/s), memory throughput (GB/s), Instruction Per Cycle (IPC), and the number of instructions compared with cuBLAS FP32 GEMM~\cite{cublas}. 
The average computation and memory throughput ratios are 2.1$\times$ and 2.2$\times$, respectively. Memory throughput is higher because our quantized GEMM writes the quantized matrix out. 
However, since GEMM is computation-intensive,
our increased computation throughput dominates the performance impacts.
Our further investigation into IPC and \# of instructions in Figure~\ref{fig:GEMM:inst} explains how {\gnn} doubles the computation throughput. Our average IPC is $\sim$70\% of the baseline, with the instruction number reduced to $\sim$31\%. Together, we can roughly double the throughput of the baseline.}

%
 
%

\begin{table}[h]
\centering
\resizebox{\columnwidth}{!}{%
\begin{tabular}{c|c|c|c|c|c}
\hline
{Dataset} & {ogbn-arxiv} & {ogbn-products} & {Pubmed} & {DBLP}   & {Amazon} \\ \hline
{Ours (GB/s)}     & {344.06}     & {491.72} & {353.38} & {331.57} & {342.93} \\ \hline
{Baseline (GB/s)} & {41.26}      & {244.22} & {131.89} & {297.67} & {105.88} \\ \hline
\end{tabular}%
}
\caption{{The achieved memory throughput using incidence-based SPMM and DGL baseline.}}
\label{tab:SPMM_profile} 
 \vspace{-.3in}
\end{table}

\begin{figure}[h]
\centering
    \subfloat[Incidence matrix optimization.]{
        \includegraphics[width=.9\columnwidth]{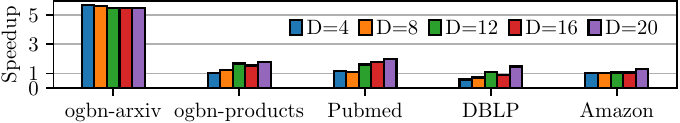}
        \label{fig:eva:incidence}
    }
    
    \subfloat[{Multi-SPMM optimization.}]{
        \includegraphics[width=.9\columnwidth]{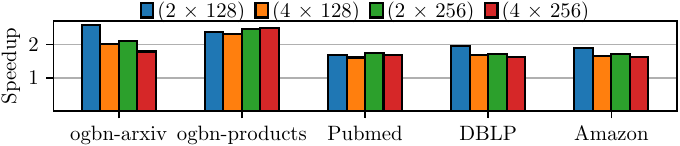}
        \label{fig:eva:SPMM_broadcast}
    }
   \caption{{\gnn} SPMM optimizations.}
  \vspace{-.1in}
\end{figure}

\textbf{SPMM.}
Figure~\ref{fig:eva:incidence} shows the performance of using incidence matrix SPMM for edge aggregation compared with DGL SPMM kernels. We set the edge features size ranging from 4 to 20. All dataset has an average $2.1\times$ speedup. The ogbn-arxiv dataset has the best speedup of $5.5\times$ on average because of the poor performance of its baseline kernel. That is, the randomness of the incidence matrix is much lower than that of the adjacency matrix from the baseline. Table~\ref{tab:SPMM_profile} shows the achieved memory throughput using our incidence-based SPMM and baseline when the feature size is 16. The irregular access for the baseline on the ogbn-arxiv dataset leads to low memory throughput. Using the incidence matrix alleviates the irregularity because the edges incidents to a node are stored adjacent in memory. 

Figure~\ref{fig:eva:SPMM_broadcast} shows the performance of using multiple SPMM's with a small edge feature dimension in a multi-head graph attention operation compared with DGL SPMM kernels. We set the node feature dimension as $(H \times D)$, and the edge features dimension as $(H\times 1)$, where $H$ represents the number of heads and $D$ represents the hidden size of each head. 
Ours has $2.1\times$, $1.9\times$, $2.0\times$, and $1.8\times$ speedup {over DGL's primitive} on average respectively for $(2\times 128)$, $(4\times 128)$, $(2\times 256)$, and $(4\times 256)$.
Increasing the number of heads leads to a smaller speedup when the hidden size is fixed because of the increased overhead of more kernel launches.
The hidden size has little impact on speedup with the same head number.

\begin{figure}[h]
 \centering
    \includegraphics[width=.8\columnwidth]{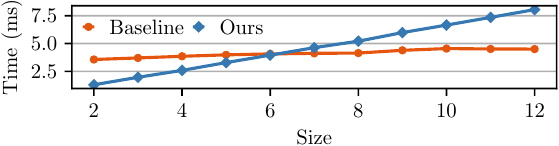}
    \caption{The performance of using multiple cuSPARSE SPMV with high edge feature dimension.}
    \label{fig:eva:SPMM_nobroadcast}
\vspace{-.15in}    
\end{figure}

Figure~\ref{fig:eva:SPMM_nobroadcast} shows the performance of using multiple cuSPARSE SpMV 
with a large edge feature dimension on ogbn-arxiv graph. We test the feature size ranging from 2 to 12. When the size is smaller than 6, ours has a $1.6\times$ speedup on average over DGL's implementation. The result shows the increased turnaround time as the number of kernels increases.

\begin{figure}[h]
 \centering
    \includegraphics[width=.9\columnwidth]{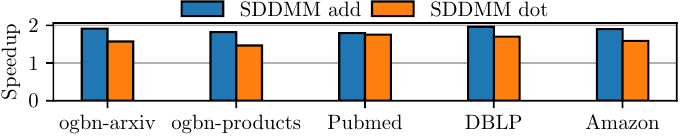}
    \caption{The performance of SDDMM operators.}
    \label{fig:eva:SDDMM}
    \vspace{-.15in}
\end{figure}
\textbf{SDDMM.}
Figure~\ref{fig:eva:SDDMM} shows the performance of quantized SDDMM compared with DGL SDDMM kernels.
We evaluate two SDDMM variants, including the row-wise dot-product (step \circled{5''} in Figure~\ref{fig:GAT_back}) and element-wise addition (step \circled{3} in Figure~\ref{fig:GAT_forward}), denoted as \textit{SDDMM dot} and \textit{SDDMM add}. The node features are matrices with the size of $(4, 64)$. 
Our SDDMM add and SDDMM dot achieves $1.9\times$ and $1.6\times$ speedups over DGL, respectively.

\begin{figure}[h]
\centering

\subfloat[{INT4 SDDMM.}]{
    \includegraphics[width=.9\columnwidth]{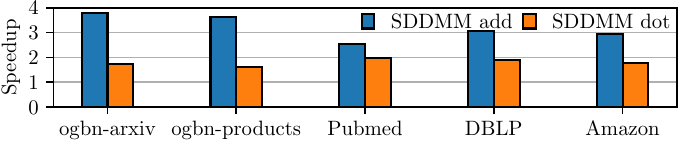}
    \label{fig:eva:int4_SDDMM}
}
\vspace{-.05in}
\subfloat[{INT4 GEMM and INT8 GEMM}.]{
    \includegraphics[width=.9\columnwidth]{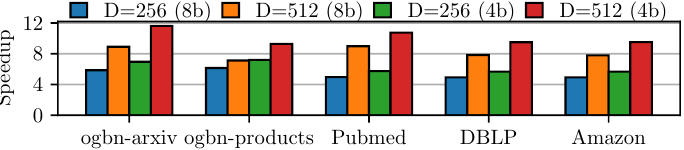}
    \label{fig:eva:int4_GEMM}
}
\vspace{-.15in}
\caption{{The turnaround time impacts by varying \# of bits.}}
\vspace{-.2in}
\end{figure}

\subsection{{Speed impact for \# of quantization bits}}

%

%
%

%
Because neither cuSPARSE nor DGL provides SPMM kernels of INT4, this section only studies INT4 GEMM and SDDMM which are implemented by {\gnn}. 
Figure~\ref{fig:eva:int4_SDDMM} shows the SDDMM performance using INT4 compared with full precision DGL primitives. The addition and dot-product kernels achieve, on average, $3.3\times$ and $1.8\times$, respectively. Dense graphs like \textit{ogbn-arxiv} and \textit{ogbn-products} enjoy more benefits from reduced memory traffic because the node embeddings are more likely to be reused by cache hit.

Figure~\ref{fig:eva:int4_GEMM}  shows the GEMM performance using INT8, and INT4 compared with cuBLAS. Note that we run the tests on an A100 GPU with INT4 hardware support. Using INT8 and INT4 leads to $5.4\times$ and $6.2\times$ average speedup when hidden size $D=256$. For $D=512$, the average speedup is $8.1\times$ and $10.1\times$, respectively. Using fewer bits shows marginal improvement because the sub-byte access under-utilizes the shared memory bandwidth.

\vspace{-.1in}
\section{Related work}\label{sec:related}

Recent years have seen a surge of efforts on GNN~\cite{wang2019dgl,DistGNN,pandey2020c,GNNLab,TileSpGEMM,9286152,bitgraphblas,deng2021low, cheshmi2022optimizing,fey2021gnnautoscale,acer2021exagraph,li2019generalized,NeuGraph,9376972,zheng2020distdgl}. {For a comprehensive study about the history and advancements in quantization for DNNs and GNNs, we refer the readers to two surveys~\cite{gholami2021survey,nagel2021white}. 
In addition to the related work in Section~\ref{sec:intr}, this section further discusses GNN primitives and inference.}

\textbf{GNN operator optimization} projects often focus on improving the SPMM and SDDMM kernels in GNN. 
GE-SpMM~\cite{gespmm} and DA-SpMM~\cite{dai2022heuristic} propose optimizations for implementing SPMM on GPU for GNN workloads. 
QGTC~\cite{QGTC} accelerates quantized GNN operations using Tensor Cores on GPU by representing the adjacency matrix as a 1-bit sparse matrix and quantizing node features in any bits, which can be computed using the 1-bit computation function on Ampere Tensor Cores.
GE-SpMM, QGTC, and DA-SpMM do not support models with multi-edge features like GAT, while {\gnn} revamps SPMM to support such models. In addition, {\gnn} also supports quantized GEMM and SDDMM. FeatGraph~\cite{FeatGraph} uses tensor compilers, providing a flexible programming interface to generate SPMM and SDDMM for various GNN operations. FeatGraph aims to exploit parallelism for customized operations between features. However, FeatGraph does not support quantization.

\textbf{GNN inference quantization} has received significant attentions recently~\cite{QGTC,tailor2021degreequant,VQGNN,SGQuant,bahri2021binary}. Unfortunately, none of these can achieve a shorter turnaround time for \textit{training} than not quantized GNN models. Particularly,~\cite{bahri2021binary} quantizes the GNN into binary with knowledge distillation to reduce accuracy loss.
Since knowledge distillation needs to train a teacher model and a student model, the training time increases. SGQuant~\cite{SGQuant} is a quantization scheme aiming to reduce memory consumption. It assigns different bits to embeddings and attention tensors on different levels, but the mismatch of datatype incurs extra conversion overhead. 
In contrast,
{\gnn} introduces a variety of framework and primitive-level system optimizations, leading to a shorter turnaround time during quantized GNN training.   

\vspace{-.1in}
\section{Conclusion}\label{sec:conclude}

{\gnn} identifies both the challenges and opportunities brought by quantization to GNN training. Particularly, {\gnn} makes the following three major contributions.
First, {\gnn} introduces various lightweight rules to maintain the accuracy for quantized GNN training. Second, we design and implement {quantization-aware} primitives and inter-primitive optimizations to reduce the turnaround time for quantized GNN training. 
Third, we integrate {\gnn} into DGL and evaluate it across a variety of GNN models and datasets to demonstrate the superior performance of {\gnn}. 




\vspace{-.1in}
\section*{Acknowledgement}
We would like to thank the anonymous reviewers for their helpful suggestions. This work was in part supported by the NSF CRII Award No. 2331536, CAREER Award No. 2326141, and NSF 2212370, 2319880, 2328948, 2319975, 2331301 and Semiconductor Research Corporation (SRC) Artificial Intelligence Hardware program. Any opinions, findings conclusions, or recommendations expressed in this material are those of the authors and do not necessarily reflect the views of the funding agencies.

{
\nocite{bengio2013estimating}
\bibliographystyle{unsrt}
\bibliography{ref}

\begin{thebibliography}{10}

\bibitem{marathe2013computational}
Madhav Marathe and Anil Kumar~S Vullikanti.
\newblock {Computational Epidemiology}.
\newblock {\em Communications of the ACM}, 56(7):88--96, 2013.

\bibitem{zhang2019circuit}
Guo Zhang, Hao He, and Dina Katabi.
\newblock {Circuit-GNN: Graph neural networks for distributed circuit design}.
\newblock In {\em {International Conference on Machine Learning}}, pages
  7364--7373. PMLR, 2019.

\bibitem{xu2017neural}
Xiaojun Xu, Chang Liu, Qian Feng, Heng Yin, Le~Song, and Dawn Song.
\newblock {Neural Network-Based Graph Embedding for Cross-Platform Binary Code
  Similarity Detection}.
\newblock In {\em Proceedings of the ACM SIGSAC Conference on Computer and
  Communications Security}, pages 363--376, 2017.

\bibitem{gaihre2021dr}
Anil Gaihre, Da~Zheng, Scott Weitze, Lingda Li, Shuaiwen~Leon Song, Caiwen
  Ding, Xiaoye~S Li, and Hang Liu.
\newblock {Dr. Top-k: Delegate-Centric Top-k on GPUs}.
\newblock In {\em Proceedings of the International Conference for High
  Performance Computing, Networking, Storage and Analysis}, pages 1--14, 2021.

\bibitem{banerjee20013}
Kaustav Banerjee, Shukri~J Souri, Pawan Kapur, and Krishna~C Saraswat.
\newblock {3-D ICs: A Novel Chip Design for Improving Deep-submicrometer
  Interconnect Performance and Systems-on-Chip Integration}.
\newblock {\em Proceedings of the IEEE}, 89(5):602--633, 2001.

\bibitem{FINGERS}
Qihang Chen, Boyu Tian, and Mingyu Gao.
\newblock {FINGERS: Exploiting Fine-Grained Parallelism in Graph Mining
  Accelerators}.
\newblock In {\em Proceedings of the 27th ACM International Conference on
  Architectural Support for Programming Languages and Operating Systems},
  ASPLOS 2022, page 43–55, New York, NY, USA, 2022.

\bibitem{chen2021re}
Shiyang Chen, Shaoyi Huang, Santosh Pandey, Bingbing Li, Guang~R Gao, Long
  Zheng, Caiwen Ding, and Hang Liu.
\newblock {ET: Re-thinking Self-Attention for Transformer Models on GPUs}.
\newblock In {\em Proceedings of the International Conference for High
  Performance Computing, Networking, Storage and Analysis}, pages 1--18, 2021.

\bibitem{gaihre2019xbfs}
Anil Gaihre, Zhenlin Wu, Fan Yao, and Hang Liu.
\newblock {XBFS: eXploring runtime optimizations for breadth-first search on
  GPUs}.
\newblock In {\em Proceedings of the 28th International symposium on
  high-performance parallel and distributed computing}, pages 121--131, 2019.

\bibitem{liu2019simd}
Hang Liu and H~Howie Huang.
\newblock {$\{$SIMD-X$\}$: Programming and processing of graph algorithms on
  $\{$GPUs$\}$}.
\newblock In {\em 2019 USENIX Annual Technical Conference (USENIX ATC 19)},
  pages 411--428, 2019.

\bibitem{GraphP}
Mingxing Zhang, Youwei Zhuo, Chao Wang, Mingyu Gao, Yongwei Wu, Kang Chen,
  Christos Kozyrakis, and Xuehai Qian.
\newblock {GraphP: Reducing Communication for PIM-Based Graph Processing with
  Efficient Data Partition}.
\newblock In {\em 2018 IEEE International Symposium on High Performance
  Computer Architecture (HPCA)}, pages 544--557, 2018.

\bibitem{lam2022graphcast}
Remi Lam, Alvaro Sanchez-Gonzalez, Matthew Willson, Peter Wirnsberger, Meire
  Fortunato, Alexander Pritzel, Suman Ravuri, Timo Ewalds, Ferran Alet, Zach
  Eaton-Rosen, et~al.
\newblock {GraphCast: Learning skillful medium-range global weather
  forecasting}.
\newblock {\em arXiv preprint arXiv:2212.12794}, 2022.

\bibitem{kipf2016semi}
Thomas~N Kipf and Max Welling.
\newblock {Semi-Supervised Classification with Graph Convolutional Networks}.
\newblock In {\em {International Conference on Learning Representations}},
  2016.

\bibitem{hamilton2017inductive}
Will Hamilton, Zhitao Ying, and Jure Leskovec.
\newblock Inductive representation learning on large graphs.
\newblock {\em {Advances in Neural Information Processing System}}, 30, 2017.

\bibitem{NIPS2013_1cecc7a7}
Antoine Bordes, Nicolas Usunier, Alberto Garcia-Duran, Jason Weston, and Oksana
  Yakhnenko.
\newblock Translating embeddings for modeling multi-relational data.
\newblock In {\em Advances in Neural Information Processing Systems},
  volume~26. Curran Associates, Inc., 2013.

\bibitem{zheng2020dgl}
Da~Zheng, Xiang Song, Chao Ma, Zeyuan Tan, Zihao Ye, Jin Dong, Hao Xiong, Zheng
  Zhang, and George Karypis.
\newblock {Dgl-ke: Training knowledge graph embeddings at scale}.
\newblock In {\em Proceedings of the 43rd International ACM SIGIR Conference on
  Research and Development in Information Retrieval}, pages 739--748, 2020.

\bibitem{InformationNidhi2021}
Nidhi Rastogi, Sharmishtha Dutta, Christian Ryan, Mohammad Zaki, Alex Gittens,
  and Charu~C. Aggarwal.
\newblock {Information Prediction using Knowledge Graphs for Contextual Malware
  Threat Intelligence}.
\newblock {\em CoRR}, abs/2102.05571, 2021.

\bibitem{xu2022payment}
Qingyu Xu, Feng Zhang, Mingde Zhang, Jidong Zhai, Bingsheng He, Cheng Yang,
  Shuhao Zhang, Jiazao Lin, Haidi Liu, and Xiaoyong Du.
\newblock {Payment behavior prediction on shared parking lots with TR-GCN}.
\newblock {\em The VLDB Journal}, pages 1--24, 2022.

\bibitem{lim2021visual}
Seung-Hwan Lim, Junghoon Chae, Guojing Cong, Drahomira Herrmannova, Robert~M
  Patton, Ramakrishnan Kannan, and Thomas~E Potok.
\newblock {Visual Understanding of COVID-19 Knowledge Graph for Predictive
  Analysis}.
\newblock In {\em 2021 IEEE International Conference on Big Data (Big Data)},
  pages 4381--4386. IEEE, 2021.

\bibitem{informatics10010008}
Yifei Wang, Shiyang Chen, Guobin Chen, Ethan Shurberg, Hang Liu, and Pengyu
  Hong.
\newblock {Motif-Based Graph Representation Learning with Application to
  Chemical Molecules}.
\newblock {\em Informatics}, 10(1), 2023.

\bibitem{won2022ulppack}
Jaeyeon Won, Jeyeon Si, Sam Son, Tae~Jun Ham, and Jae~W Lee.
\newblock {ULPPACK: Fast Sub-8-bit Matrix Multiply on Commodity SIMD Hardware}.
\newblock {\em Proceedings of Machine Learning and Systems}, 4:52--63, 2022.

\bibitem{GraSP}
Minjia Zhang, Wenhan Wang, and Yuxiong He.
\newblock {GraSP: Optimizing Graph-Based Nearest Neighbor Search with Subgraph
  Sampling and Pruning}.
\newblock In {\em Proceedings of the Fifteenth ACM International Conference on
  Web Search and Data Mining}, WSDM '22, page 1395–1405, New York, NY, USA,
  2022. ACM.

\bibitem{GAT2018graph}
Petar Veličković, Guillem Cucurull, Arantxa Casanova, Adriana Romero, Pietro
  Liò, and Yoshua Bengio.
\newblock {Graph Attention Networks}.
\newblock In {\em International Conference on Learning Representations}, 2018.

\bibitem{tensorcore}
Nvidia.
\newblock {Tensor Cores}.
\newblock Retrieved from \url{https://developer.nvidia.com/tensor-cores}.
\newblock Accessed: 2022, Nov 26.

\bibitem{wang2018training}
Naigang Wang, Jungwook Choi, Daniel Brand, Chia-Yu Chen, and Kailash
  Gopalakrishnan.
\newblock {Training Deep Neural Networks with 8-bit Floating Point Numbers}.
\newblock {\em Advances in Neural Information Processing Systems}, 31, 2018.

\bibitem{koster2017flexpoint}
Urs K{\"o}ster, Tristan Webb, Xin Wang, Marcel Nassar, Arjun~K Bansal, William
  Constable, Oguz Elibol, Scott Gray, Stewart Hall, Luke Hornof, et~al.
\newblock {Flexpoint: An Adaptive Numerical Format for Efficient Training of
  Deep Neural Networks}.
\newblock {\em Advances in Neural Information Processing Systems}, 30, 2017.

\bibitem{das2018mixed}
Dipankar Das, Naveen Mellempudi, Dheevatsa Mudigere, Dhiraj Kalamkar, Sasikanth
  Avancha, Kunal Banerjee, Srinivas Sridharan, Karthik Vaidyanathan, Bharat
  Kaul, Evangelos Georganas, et~al.
\newblock {Mixed Precision Training of Convolutional Neural Networks using
  Integer Operations}.
\newblock In {\em International Conference on Learning Representations}, 2018.

\bibitem{sakr2018per}
Charbel Sakr and Naresh Shanbhag.
\newblock {Per-Tensor Fixed-Point Quantization of the Back-Propagation
  Algorithm}.
\newblock In {\em International Conference on Learning Representations}, 2018.

\bibitem{yang2019swalp}
Guandao Yang, Tianyi Zhang, Polina Kirichenko, Junwen Bai, Andrew~Gordon
  Wilson, and Chris De~Sa.
\newblock {SWALP: Stochastic Weight Averaging in Low Precision Training}.
\newblock In {\em International Conference on Machine Learning}, pages
  7015--7024. PMLR, 2019.

\bibitem{zhu2020towards}
Feng Zhu, Ruihao Gong, Fengwei Yu, Xianglong Liu, Yanfei Wang, Zhelong Li,
  Xiuqi Yang, and Junjie Yan.
\newblock {Towards Unified INT8 Training for Convolutional Neural Network}.
\newblock In {\em Proceedings of the IEEE/CVF Conference on Computer Vision and
  Pattern Recognition}, pages 1969--1979, 2020.

\bibitem{chen2021actnn}
Jianfei Chen, Lianmin Zheng, Zhewei Yao, Dequan Wang, Ion Stoica, Michael
  Mahoney, and Joseph Gonzalez.
\newblock {ActNN: Reducing Training Memory Footprint via 2-Bit Activation
  Compressed Training}.
\newblock In {\em Proceedings of the 38th International Conference on Machine
  Learning}, volume 139 of {\em Proceedings of Machine Learning Research},
  pages 1803--1813. PMLR, 18--24 Jul 2021.

\bibitem{chen2022tinykg}
Huiyuan Chen, Xiaoting Li, Kaixiong Zhou, Xia Hu, Chin-Chia~Michael Yeh, Yan
  Zheng, and Hao Yang.
\newblock {TinyKG: Memory-Efficient Training Framework for Knowledge Graph
  Neural Recommender Systems}.
\newblock In {\em Proceedings of the 16th ACM Conference on Recommender
  Systems}, pages 257--267, 2022.

\bibitem{liu2022exact}
Zirui Liu, Kaixiong Zhou, Fan Yang, Li~Li, Rui Chen, and Xia Hu.
\newblock {EXACT: Scalable graph neural networks training via extreme
  activation compression}.
\newblock In {\em International Conference on Learning Representations}, 2022.

\bibitem{liu2022gact}
Xiaoxuan Liu, Lianmin Zheng, Dequan Wang, Yukuo Cen, Weize Chen, Xu~Han,
  Jianfei Chen, Zhiyuan Liu, Jie Tang, Joey Gonzalez, et~al.
\newblock {GACT: Activation Compressed Training for Generic Network
  Architectures}.
\newblock {\em International Conference on Machine Learning}, 2022.

\bibitem{evans2021ac}
R~David Evans and Tor Aamodt.
\newblock {AC-GC: Lossy activation compression with guaranteed convergence}.
\newblock {\em Advances in Neural Information Processing Systems},
  34:27434--27448, 2021.

\bibitem{tailor2021degreequant}
Shyam~Anil Tailor, Javier Fernandez-Marques, and Nicholas~Donald Lane.
\newblock {Degree-Quant: Quantization-Aware Training for Graph Neural
  Networks}.
\newblock In {\em International Conference on Learning Representations}, 2021.

\bibitem{zafrir2019q8bert}
Ofir Zafrir, Guy Boudoukh, Peter Izsak, and Moshe Wasserblat.
\newblock {Q8BERT: Quantized 8bit BERT}.
\newblock In {\em 2019 Fifth Workshop on Energy Efficient Machine Learning and
  Cognitive Computing-NeurIPS Edition (EMC2-NIPS)}, pages 36--39. IEEE, 2019.

\bibitem{bhandare2019efficient}
Aishwarya Bhandare, Vamsi Sripathi, Deepthi Karkada, Vivek Menon, Sun Choi,
  Kushal Datta, and Vikram Saletore.
\newblock {Efficient 8-Bit Quantization of Transformer Neural Machine Language
  Translation Model}.
\newblock {\em arXiv preprint arXiv:1906.00532}, 2019.

\bibitem{jin2020adabits}
Qing Jin, Linjie Yang, and Zhenyu Liao.
\newblock {Adabits: Neural Network Quantization with Adaptive Bit-Widths}.
\newblock In {\em Proceedings of the IEEE/CVF Conference on Computer Vision and
  Pattern Recognition}, pages 2146--2156, 2020.

\bibitem{dong2019hawq}
Zhen Dong, Zhewei Yao, Amir Gholami, Michael~W Mahoney, and Kurt Keutzer.
\newblock {HAWQ: Hessian AWare Quantization of Neural Networks with
  Mixed-Precision}.
\newblock In {\em Proceedings of the IEEE/CVF International Conference on
  Computer Vision}, pages 293--302, 2019.

\bibitem{esser2015backpropagation}
Steve~K Esser, Rathinakumar Appuswamy, Paul Merolla, John~V Arthur, and
  Dharmendra~S Modha.
\newblock {Backpropagation for Energy-Efficient Neuromorphic Computing}.
\newblock {\em Advances in Neural Information Processing Systems}, 28, 2015.

\bibitem{courbariaux2015binaryconnect}
Matthieu Courbariaux, Yoshua Bengio, and Jean-Pierre David.
\newblock {Binaryconnect: Training deep neural networks with binary weights
  during propagations}.
\newblock {\em Advances in Neural Information Processing Systems}, 28, 2015.

\bibitem{jacob2018quantization}
Benoit Jacob, Skirmantas Kligys, Bo~Chen, Menglong Zhu, Matthew Tang, Andrew
  Howard, Hartwig Adam, and Dmitry Kalenichenko.
\newblock Quantization and training of neural networks for efficient
  integer-arithmetic-only inference.
\newblock In {\em Proceedings of the IEEE Conference on Computer Vision and
  Pattern Recognition (CVPR)}, June 2018.

\bibitem{RGCN}
Michael Schlichtkrull, Thomas~N. Kipf, Peter Bloem, Rianne van~den Berg, Ivan
  Titov, and Max Welling.
\newblock {Modeling Relational Data with Graph Convolutional Networks}.
\newblock In {\em The Semantic Web}, pages 593--607, Cham, 2018. Springer
  International Publishing.

\bibitem{7536145}
Chuan Shi, Yitong Li, Jiawei Zhang, Yizhou Sun, and Philip~S. Yu.
\newblock A survey of heterogeneous information network analysis.
\newblock {\em IEEE Transactions on Knowledge and Data Engineering},
  29(1):17--37, 2017.

\bibitem{hu2020heterogeneous}
Ziniu Hu, Yuxiao Dong, Kuansan Wang, and Yizhou Sun.
\newblock {Heterogeneous Graph Transformer}.
\newblock In {\em Proceedings of The Web Conference 2020}, pages 2704--2710,
  2020.

\bibitem{gupta2015deep}
Suyog Gupta, Ankur Agrawal, Kailash Gopalakrishnan, and Pritish Narayanan.
\newblock {Deep Learning with Limited Numerical Precision}.
\newblock In {\em International Conference on Machine Learning}, pages
  1737--1746. PMLR, 2015.

\bibitem{cuRAND}
CUDA Nvidia.
\newblock {cuRAND library programming guide}.
\newblock {\em NVIDIA Corporation. edit}, 1, 2022.

\bibitem{blackman2021scrambled}
David Blackman and Sebastiano Vigna.
\newblock {Scrambled Linear Pseudorandom Number Generators}.
\newblock {\em ACM Transactions on Mathematical Software (TOMS)}, 47(4):1--32,
  2021.

\bibitem{abboud2020surprising}
Ralph Abboud, İsmail~İlkan Ceylan, Martin Grohe, and Thomas Lukasiewicz.
\newblock {The Surprising Power of Graph Neural Networks with Random Node
  Initialization}.
\newblock In {\em Proceedings of the Thirtieth International Joint Conference
  on Artificial Intelligence, {IJCAI-21}}, pages 2112--2118, 8 2021.

\bibitem{park2018training}
Hyunsun Park, Jun~Haeng Lee, Youngmin Oh, Sangwon Ha, and Seungwon Lee.
\newblock {Training Deep Neural Network in Limited Precision}.
\newblock {\em arXiv preprint arXiv:1810.05486}, 2018.

\bibitem{chen2017fxpnet}
Xi~Chen, Xiaolin Hu, Hucheng Zhou, and Ningyi Xu.
\newblock {FxpNet: Training a deep convolutional neural network in fixed-point
  representation}.
\newblock In {\em 2017 International Joint Conference on Neural Networks
  (IJCNN)}, pages 2494--2501. IEEE, 2017.

\bibitem{zhou2016dorefa}
Shuchang Zhou, Yuxin Wu, Zekun Ni, Xinyu Zhou, He~Wen, and Yuheng Zou.
\newblock {DoReFa-Net: Training Low Bitwidth Convolutional Neural Networks with
  Low Bitwidth Gradients}.
\newblock {\em arXiv preprint arXiv:1606.06160}, 2016.

\bibitem{chen2018learning}
Tianqi Chen, Lianmin Zheng, Eddie Yan, Ziheng Jiang, Thierry Moreau, Luis Ceze,
  Carlos Guestrin, and Arvind Krishnamurthy.
\newblock Learning to optimize tensor programs.
\newblock {\em Advances in Neural Information Processing Systems}, 31, 2018.

\bibitem{zheng2020ansor}
Lianmin Zheng, Chengfan Jia, Minmin Sun, Zhao Wu, Cody~Hao Yu, Ameer Haj-Ali,
  Yida Wang, Jun Yang, Danyang Zhuo, Koushik Sen, et~al.
\newblock {Ansor: Generating High-performance Tensor Programs for Deep
  Learning}.
\newblock In {\em Proceedings of the 14th USENIX Conference on Operating
  Systems Design and Implementation}, pages 863--879, 2020.

\bibitem{tillet2019triton}
Philippe Tillet, Hsiang-Tsung Kung, and David Cox.
\newblock {Triton: An Intermediate Language and Compiler for Tiled Neural
  Network Computations}.
\newblock In {\em Proceedings of the 3rd ACM SIGPLAN International Workshop on
  Machine Learning and Programming Languages}, pages 10--19, 2019.

\bibitem{wang2020microsoft}
Kuansan Wang, Zhihong Shen, Chiyuan Huang, Chieh-Han Wu, Yuxiao Dong, and
  Anshul Kanakia.
\newblock {Microsoft Academic Graph: When experts are not enough}.
\newblock {\em Quantitative Science Studies}, 1(1):396--413, 2020.

\bibitem{namata2012query}
Galileo Namata, Ben London, Lise Getoor, Bert Huang, and UMD EDU.
\newblock {Query-driven Active Surveying for Collective Classification}.
\newblock In {\em 10th International Workshop on Mining and Learning with
  Graphs}, volume~8, page~1, 2012.

\bibitem{Bhatia16}
K.~Bhatia, K.~Dahiya, H.~Jain, P.~Kar, A.~Mittal, Y.~Prabhu, and M.~Varma.
\newblock {The Extreme Classification Repository: Multi-label Datasets and
  code}, 2016.

\bibitem{yang2015defining}
Jaewon Yang and Jure Leskovec.
\newblock {Defining and Evaluating Network Communities based on Ground-truth}.
\newblock {\em Knowledge and Information Systems}, 42(1):181--213, 2015.

\bibitem{leskovec2007dynamics}
Jure Leskovec, Lada~A Adamic, and Bernardo~A Huberman.
\newblock {The Dynamics of Viral Marketing}.
\newblock {\em ACM Transactions on the Web (TWEB)}, 1(1):5--es, 2007.

\bibitem{cusparse}
M~Naumov, LS~Chien, P~Vandermersch, and U~Kapasi.
\newblock Cusparse library.
\newblock In {\em GPU Technology Conference (GTC)}, 2010.

\bibitem{pytorch}
Adam Paszke, Sam Gross, Francisco Massa, Adam Lerer, James Bradbury, Gregory
  Chanan, Trevor Killeen, Zeming Lin, Natalia Gimelshein, Luca Antiga, et~al.
\newblock {PyTorch: An imperative style, high-performance deep learning
  library}.
\newblock In {\em Advances in Neural Information Processing Systems}, pages
  8024--8035, 2019.

\bibitem{wang2019dgl}
Minjie Wang, Da~Zheng, Zihao Ye, Quan Gan, Mufei Li, Xiang Song, Jinjing Zhou,
  Chao Ma, Lingfan Yu, Yu~Gai, Tianjun Xiao, Tong He, George Karypis, Jinyang
  Li, and Zheng Zhang.
\newblock {Deep Graph Library: A Graph-Centric, Highly-Performant Package for
  Graph Neural Networks}.
\newblock {\em arXiv preprint arXiv:1909.01315}, 2019.

\bibitem{cublas}
CUDA Nvidia.
\newblock {cuBLAS library programming guide}.
\newblock {\em NVIDIA Corporation. edit}, 1, 2007.

\bibitem{DistGNN}
Vasimuddin Md, Sanchit Misra, Guixiang Ma, Ramanarayan Mohanty, Evangelos
  Georganas, Alexander Heinecke, Dhiraj Kalamkar, Nesreen~K. Ahmed, and
  Sasikanth Avancha.
\newblock {DistGNN: Scalable Distributed Training for Large-Scale Graph Neural
  Networks}.
\newblock In {\em Proceedings of the International Conference for High
  Performance Computing, Networking, Storage and Analysis}, SC '21, New York,
  NY, USA, 2021. ACM.

\bibitem{pandey2020c}
Santosh Pandey, Lingda Li, Adolfy Hoisie, Xiaoye~S Li, and Hang Liu.
\newblock {C-SAW: A framework for graph sampling and random walk on GPUs}.
\newblock In {\em SC20: International Conference for High Performance
  Computing, Networking, Storage and Analysis}, pages 1--15. IEEE, 2020.

\bibitem{GNNLab}
Jianbang Yang, Dahai Tang, Xiaoniu Song, Lei Wang, Qiang Yin, Rong Chen,
  Wenyuan Yu, and Jingren Zhou.
\newblock {GNNLab: A Factored System for Sample-Based GNN Training over GPUs}.
\newblock In {\em Proceedings of the Seventeenth European Conference on
  Computer Systems}, EuroSys '22, page 417–434, New York, NY, USA, 2022. ACM.

\bibitem{TileSpGEMM}
Yuyao Niu, Zhengyang Lu, Haonan Ji, Shuhui Song, Zhou Jin, and Weifeng Liu.
\newblock {TileSpGEMM: A Tiled Algorithm for Parallel Sparse General
  Matrix-Matrix Multiplication on GPUs}.
\newblock In {\em Proceedings of the 27th ACM SIGPLAN Symposium on Principles
  and Practice of Parallel Programming}, PPoPP '22, page 90–106, New York,
  NY, USA, 2022. ACM.

\bibitem{9286152}
Jesun~Sahariar Firoz, Ang Li, Jiajia Li, and Kevin Barker.
\newblock {On the Feasibility of Using Reduced-Precision Tensor Core Operations
  for Graph Analytics}.
\newblock In {\em 2020 IEEE High Performance Extreme Computing Conference
  (HPEC)}, pages 1--7, 2020.

\bibitem{bitgraphblas}
Jou-An Chen, Hsin-Hsuan Sung, Xipeng Shen, Nathan Tallent, Kevin Barker, and
  Ang Li.
\newblock {Bit-GraphBLAS: Bit-Level Optimizations of Matrix-Centric Graph
  Processing on GPU}, 2022.

\bibitem{deng2021low}
Zhaoxia Deng, Jongsoo Park, Ping Tak~Peter Tang, Haixin Liu, Jie Yang, Hector
  Yuen, Jianyu Huang, Daya Khudia, Xiaohan Wei, Ellie Wen, et~al.
\newblock {Low-Precision Hardware Architectures Meet Recommendation Model
  Inference at Scale}.
\newblock {\em IEEE Micro}, 41(5):93--100, 2021.

\bibitem{cheshmi2022optimizing}
Kazem Cheshmi, Michelle~Mills Strout, and Maryam~Mehri Dehnavi.
\newblock {Optimizing Sparse Computations Jointly}.
\newblock In {\em Proceedings of the 27th ACM SIGPLAN Symposium on Principles
  and Practice of Parallel Programming}, pages 459--460, 2022.

\bibitem{fey2021gnnautoscale}
Matthias Fey, Jan~E Lenssen, Frank Weichert, and Jure Leskovec.
\newblock {GNNAutoScale: Scalable and Expressive Graph Neural Networks via
  Historical Embeddings}.
\newblock In {\em International conference on machine learning}, pages
  3294--3304. PMLR, 2021.

\bibitem{acer2021exagraph}
Seher Acer, Ariful Azad, Erik~G Boman, Ayd{\i}n Bulu{\c{c}}, Karen~D Devine,
  SM~Ferdous, Nitin Gawande, Sayan Ghosh, Mahantesh Halappanavar, Ananth
  Kalyanaraman, et~al.
\newblock {EXAGRAPH: Graph and combinatorial methods for enabling exascale
  applications}.
\newblock {\em The International Journal of High Performance Computing
  Applications}, 35(6):553--571, 2021.

\bibitem{li2019generalized}
Jiayu Li, Fugang Wang, Takuya Araki, and Judy Qiu.
\newblock {Generalized Sparse Matrix-Matrix Multiplication for Vector Engines
  and Graph Applications}.
\newblock In {\em 2019 IEEE/ACM Workshop on Memory Centric High Performance
  Computing (MCHPC)}, pages 33--42. IEEE, 2019.

\bibitem{NeuGraph}
Lingxiao Ma, Zhi Yang, Youshan Miao, Jilong Xue, Ming Wu, Lidong Zhou, and
  Yafei Dai.
\newblock {{NeuGraph}: Parallel Deep Neural Network Computation on Large
  Graphs}.
\newblock In {\em 2019 USENIX Annual Technical Conference}, pages 443--458,
  Renton, WA, July 2019. USENIX Association.

\bibitem{9376972}
Youhui Bai, Cheng Li, Zhiqi Lin, Yufei Wu, Youshan Miao, Yunxin Liu, and
  Yinlong Xu.
\newblock {Efficient Data Loader for Fast Sampling-Based GNN Training on Large
  Graphs}.
\newblock {\em IEEE Transactions on Parallel and Distributed Systems},
  32(10):2541--2556, 2021.

\bibitem{zheng2020distdgl}
Da~Zheng, Chao Ma, Minjie Wang, Jinjing Zhou, Qidong Su, Xiang Song, Quan Gan,
  Zheng Zhang, and George Karypis.
\newblock {DistDGL: Distributed Graph Neural Network Training for Billion-scale
  Graphs}.
\newblock In {\em 2020 IEEE/ACM 10th Workshop on Irregular Applications:
  Architectures and Algorithms (IA3)}, pages 36--44. IEEE, 2020.

\bibitem{gholami2021survey}
Amir Gholami, Sehoon Kim, Zhen Dong, Zhewei Yao, Michael~W Mahoney, and Kurt
  Keutzer.
\newblock {A Survey of Quantization Methods for Efficient Neural Network
  Inference}.
\newblock {\em arXiv preprint arXiv:2103.13630}, 2021.

\bibitem{nagel2021white}
Markus Nagel, Marios Fournarakis, Rana~Ali Amjad, Yelysei Bondarenko, Mart van
  Baalen, and Tijmen Blankevoort.
\newblock {A White Paper on Neural Network Quantization}.
\newblock {\em arXiv preprint arXiv:2106.08295}, 2021.

\bibitem{gespmm}
Guyue Huang, Guohao Dai, Yu~Wang, and Huazhong Yang.
\newblock {GE-SpMM: General-Purpose Sparse Matrix-Matrix Multiplication on GPUs
  for Graph Neural Networks}.
\newblock In {\em SC20: International Conference for High Performance
  Computing, Networking, Storage and Analysis}, pages 1--12, 2020.

\bibitem{dai2022heuristic}
Guohao Dai, Guyue Huang, Shang Yang, Zhongming Yu, Hengrui Zhang, Yufei Ding,
  Yuan Xie, Huazhong Yang, and Yu~Wang.
\newblock {Heuristic Adaptability to Input Dynamics for SpMM on GPUs }.
\newblock In {\em Proceedings of the 59th ACM/IEEE Design Automation
  Conference}, pages 595--600, 2022.

\bibitem{QGTC}
Yuke Wang, Boyuan Feng, and Yufei Ding.
\newblock {{QGTC}: Accelerating Quantized Graph Neural Networks via GPU Tensor
  Core}.
\newblock In {\em Proceedings of the 27th ACM SIGPLAN Symposium on Principles
  and Practice of Parallel Programming}, PPoPP '22, page 107–119, New York,
  NY, USA, 2022. ACM.

\bibitem{FeatGraph}
Yuwei Hu, Zihao Ye, Minjie Wang, Jiali Yu, Da~Zheng, Mu~Li, Zheng Zhang, Zhiru
  Zhang, and Yida Wang.
\newblock {FeatGraph: A Flexible and Efficient Backend for Graph Neural Network
  Systems}.
\newblock In {\em Proceedings of the International Conference for High
  Performance Computing, Networking, Storage and Analysis}, SC. IEEE, 2020.

\bibitem{VQGNN}
Mucong Ding, Kezhi Kong, Jingling Li, Chen Zhu, John Dickerson, Furong Huang,
  and Tom Goldstein.
\newblock {VQ-GNN}: A universal framework to scale up graph neural networks
  using vector quantization.
\newblock In {\em Advances in Neural Information Processing Systems},
  volume~34, pages 6733--6746. Curran Associates, Inc., 2021.

\bibitem{SGQuant}
Boyuan Feng, Yuke Wang, Xu~Li, Shu Yang, Xueqiao Peng, and Yufei Ding.
\newblock {SGQuant: Squeezing the Last Bit on Graph Neural Networks with
  Specialized Quantization}.
\newblock In {\em 2020 IEEE 32nd International Conference on Tools with
  Artificial Intelligence (ICTAI)}, pages 1044--1052, 2020.

\bibitem{bahri2021binary}
Mehdi Bahri, Ga{\'e}tan Bahl, and Stefanos Zafeiriou.
\newblock {Binary Graph Neural Networks}.
\newblock In {\em Proceedings of the IEEE/CVF Conference on Computer Vision and
  Pattern Recognition}, pages 9492--9501, 2021.

\bibitem{bengio2013estimating}
Yoshua Bengio, Nicholas L{\'e}onard, and Aaron Courville.
\newblock {Estimating or Propagating Gradients Through Stochastic Neurons for
  Conditional Computation}.
\newblock {\em arXiv preprint arXiv:1308.3432}, 2013.

\end{thebibliography}
}

\end{document}